\documentclass[journal]{IEEEtran}

\usepackage{amsmath,amsfonts,amssymb}

\usepackage{algorithmic}
\usepackage{algorithm}
\usepackage{array}
\usepackage[caption=false,font=normalsize,labelfont=sf,textfont=rm]{subfig}
\usepackage{textcomp}
\usepackage{stfloats}
\usepackage{url}
\usepackage{verbatim}
\usepackage{graphicx}

\graphicspath{{../}{./}}

\usepackage{multirow}
\usepackage{cite}
\makeatletter
\def\ps@IEEEtitlepagestyle{%
  \def\@oddhead{\hbox{}\scriptsize PAPER FOR IEEE GRSM 2026\hfill\thepage}%
  \def\@evenhead{\scriptsize PAPER FOR IEEE GRSM 2026\hfill\thepage}%
  \def\@oddfoot{}%
  \def\@evenfoot{}%
}
\def\ps@headings{%
  \def\@oddhead{\hbox{}\scriptsize PAPER FOR IEEE GRSM 2026\hfill\thepage}%
  \def\@evenhead{\scriptsize PAPER FOR IEEE GRSM 2026\hfill\thepage}%
  \def\@oddfoot{}%
  \def\@evenfoot{}%
}
\def\hlinew#1{%
  \noalign{\ifnum0=`}\fi\hrule \@height #1 \futurelet
   \reserved@a\@xhline}
\makeatother

\newcommand{\R}{\mathbb{R}}
\newcommand{\syn}{\mathrm{syn}}
\newcommand{\ms}{\mathrm{ms}}
\newcommand{\hsi}{\mathrm{hsi}}

\usepackage{xcolor}
\newcommand{\highlight}[1]{{#1}}

\begin{document}

\title{HyperSAM: A Promptable Foundation Model for Hyperspectral Remote Sensing}


\author{Li~Pang,
Xinqiao~Wu,
Jing~Yao,~\IEEEmembership{Senior Member,~IEEE},
Pedram~Ghamisi,~\IEEEmembership{Senior Member,~IEEE},
Jun~Zhou,~\IEEEmembership{Fellow,~IEEE},
Zhengchao~Chen,
Deyu~Meng, and
Xiangyong~Cao,~\IEEEmembership{Senior Member,~IEEE}%
\thanks{This work is supported by Fundamental and Interdisciplinary Disciplines Break through Plan of the Ministry of Education of China (No. JYB2025XDXM101), China NSFC Projects under Contract 62272375 and Contract 42671469, and Tianyuan Fund for Mathematics of the National Natural Science Foundation of China (Grant No. 12426105). }
\thanks{Li Pang and Deyu Meng are with the School of Mathematics and Statistics, Xi'an Jiaotong University, Xi'an 710049, China (e-mail: 2195112306@stu.xjtu.edu.cn; dymeng@mail.xjtu.edu.cn).}%
\thanks{Xinqiao Wu is with the Faculty of Electronic and Information Engineering, Xi'an Jiaotong University, Xi'an 710049, China (e-mail: 2232112661@stu.xjtu.edu.cn).}%
\thanks{Jing Yao and Zhengchao Chen are with the State Key Laboratory of Remote Sensing and Digital Earth, Aerospace Information Research Institute, Chinese Academy of Sciences, Beijing 100094, China (e-mail: yaojing@aircas.ac.cn; chenzc@aircas.ac.cn).}%
\thanks{Pedram Ghamisi is with Helmholtz-Zentrum Dresden-Rossendorf, 09599 Freiberg, Germany, and also with Faculty of Electrical and Computer Engineering, University of Iceland, 101 Reykjavik, Iceland (e-mail: p.ghamisi@gmail.com).}%
\thanks{Jun Zhou is with the School of Information and Communication Technology, Griffith University, Nathan, QLD 4111, Australia (e-mail: jun.zhou@griffith.edu.au).}%
\thanks{Xiangyong Cao is with the School of Computer Science and Technology, Xi'an Jiaotong University, Xi'an 710049, China (e-mail: caoxiangyong@mail.xjtu.edu.cn).}%
\thanks{Corresponding authors: Jing Yao; Xiangyong Cao.}
}

\markboth{PAPER FOR IEEE GRSM 2026}{PAPER FOR IEEE GRSM 2026}

\maketitle
\thispagestyle{IEEEtitlepagestyle}

\begin{abstract}
Hyperspectral remote sensing provides dense spectral measurements that are indispensable for material-level Earth observation, yet the construction of a general-purpose hyperspectral foundation model remains difficult. Two bottlenecks are especially limiting. First, large hyperspectral corpora rarely provide high spatial resolution together with reliable dense annotations. Second, many hyperspectral models are still trained almost from scratch, so the geometric and interactive priors learned by modern vision foundation models are not fully reused. To alleviate these issues, we \highlight{present} \textbf{HyperSAM}, a promptable hyperspectral foundation model that couples a data-centric hyperspectral synthesis pipeline with a spectral adaptation architecture based on Segment Anything Model 3 (SAM3). On the data side, HyperSAM synthesizes full-spectrum hyperspectral cubes from high-resolution SpaceNet multispectral imagery through a physics-informed abundance-transfer generator, while SAM3-derived pseudo-masks provide object-centric supervision. On the model side, the latest implementation uses a frozen SAM3 RGB image branch, a trainable hyperspectral side encoder initialized from the RGB vision transformer (ViT), ControlNet-style zero-initialized feature injection, and a lightweight mixture-of-experts mask refiner. To enhance training robustness against noisy pseudo-labels, Cross-modal Sample Selection (CromSS)-style confidence selection is incorporated for noisy-label weighting. Extensive experiments show that HyperSAM obtains strong generalization on diverse hyperspectral tasks (e.g., classification, anomaly detection, change detection, target detection, and airborne oil-spill mapping) and that high-quality synthetic hyperspectral data can be more effective than simply scaling noisy hyperspectral supervision.
\end{abstract}


\section{Introduction}
\label{sec:introduction}

Hyperspectral imagery (HSI) records hundreds of contiguous spectral bands for every spatial location. Compared with RGB or conventional multispectral imagery, the resulting data cube preserves fine absorption and reflection patterns that reveal material composition, vegetation status, mineral signatures, man-made objects, and subtle surface changes. These properties make hyperspectral analysis central to land-cover classification, anomaly discovery, target detection, crop monitoring, mineral exploration, urban interpretation, and disaster assessment~\cite{plaza2009recent,ghamisi2017advances,paoletti2019deep,reed1990adaptive,manolakis2001hyperspectral,wang2024hypersigma,li2025hyperfree,pang2026special}. However, hyperspectral remote sensing also creates a difficult learning problem: the data are high dimensional, sensor-dependent, costly to annotate, and often acquired at modest spatial resolution. A model that performs well on one scene or task can easily fail when the wavelength range, ground sampling distance, object scale, or annotation protocol changes~\cite{chen2025cangling}.

Foundation models have changed the way many visual recognition problems are solved. Instead of training a task-specific model from a small labeled dataset, a general model is first trained on a broad corpus and then adapted or prompted for downstream use. In remote sensing, recent work has explored large-scale pretraining for multispectral and hyperspectral data, including SatMAE, SpectralGPT, DOFA, HyperSIGMA, and HyperFree~\cite{cong2022satmae,hong2024spectralgpt,xiong2024dofa,wang2024hypersigma,li2025hyperfree}. These models demonstrate that pretraining can improve transferability, but the hyperspectral setting still lags behind the rapid progress of natural-image foundation models such as the Segment Anything Model (SAM), SAM2, and SAM3~\cite{kirillov2023sam,ravi2025sam2,carion2025sam3segmentconcepts}. The gap is not only architectural. It is also caused by the absence of a hyperspectral counterpart to the massive, diverse, and accurately annotated data used to train generic segmentation models\highlight{.}

\highlight{From the perspective of downstream use, remote-sensing foundation models follow two distinct paradigms. Representation-learning models such as SatMAE, SpectralGPT, DOFA, HyperSIGMA, and SpectralEarth, as well as generic self-supervised models such as DINOv3, transfer a pretrained encoder but generally require a task-specific head, linear probe, or fine-tuning procedure for a new scene or task~\cite{cong2022satmae,hong2024spectralgpt,xiong2024dofa,wang2024hypersigma,braham2025spectralearth, simeoni2025dinov3}. Promptable models instead expose a prompt-to-mask interface and can reuse a frozen checkpoint without retraining the backbone. HyperFree is the closest prior promptable hyperspectral model, whereas HyperSAM targets the same tuning-free setting while preserving a frozen SAM3 spatial prior and injecting full-spectrum hyperspectral evidence~\cite{li2025hyperfree,carion2025sam3segmentconcepts}.}
 \begin{figure}[t]
    \centering
    \includegraphics[width=1\linewidth]{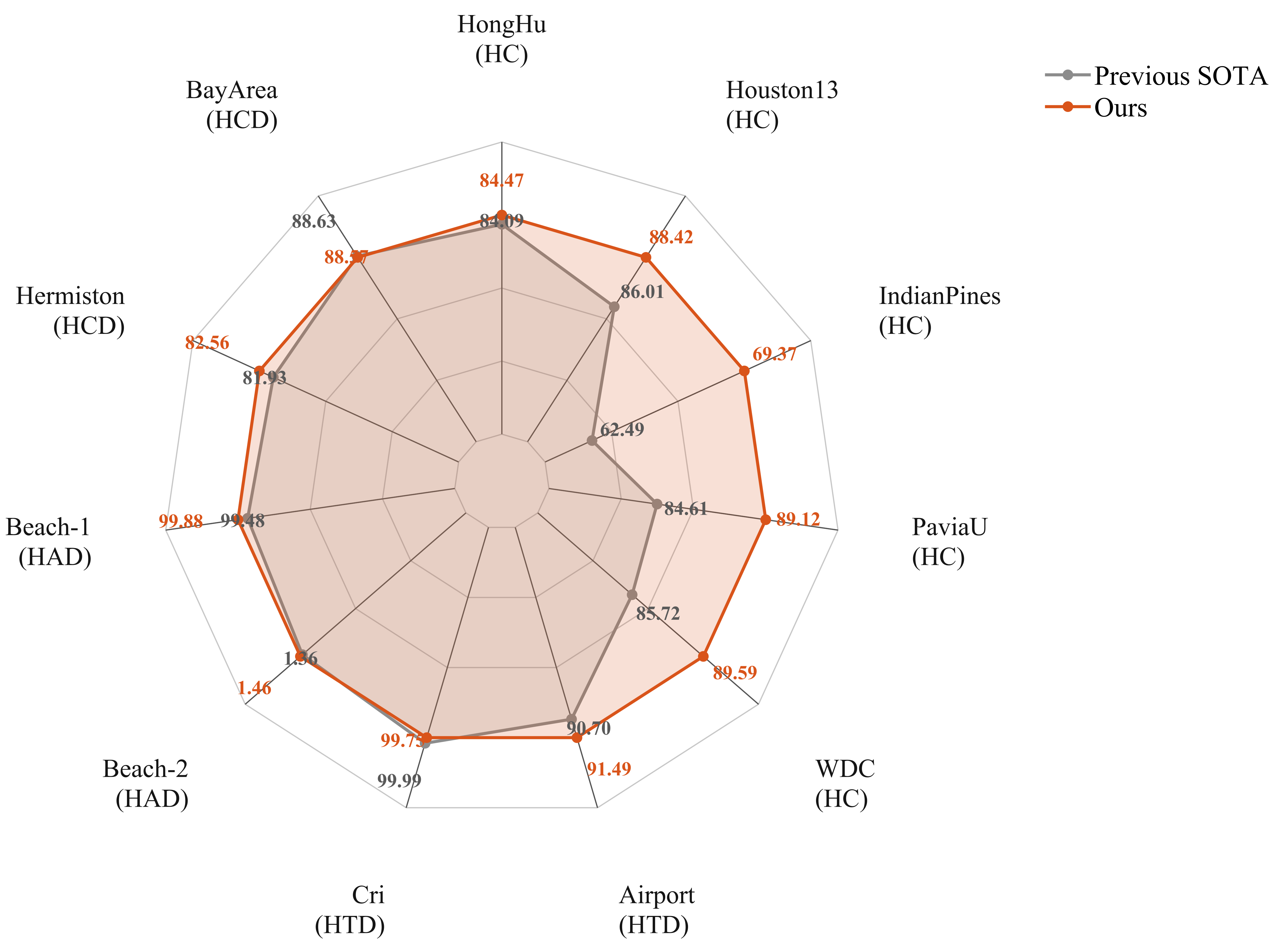}
    \caption{Cross-task performance comparison between HyperSAM and representative baselines. Results are normalized from hyperspectral classification (HC), hyperspectral anomaly detection (HAD), hyperspectral change detection (HCD), and hyperspectral target detection (HTD) metrics.}
    \label{fig:radar}
\end{figure}
We identify two bottlenecks that typically appear when building hyperspectral foundation models:

\textbf{Data quality bottleneck.} Classical hyperspectral benchmarks, such as Indian Pines and Pavia University (PaviaU), are valuable because they contain clean annotations and well-studied protocols~\cite{baumgardner2015indianpines,gic2021paviau}. Yet they are too small and homogeneous to support foundation-level learning. Larger resources, including HySpecNet-11k and HyperGlobal-style corpora, improve scale but often trade away either spatial detail or object-level labels~\cite{fuchs2023hyspecnet,mani2025ohid1,wang2024hypersigma}. HyperFree makes a notable step toward promptable hyperspectral foundation modeling by constructing the Hyper-Seg corpus~\cite{li2025hyperfree}. Its data engine generates pseudo-masks by applying SAM-H to selected three-channel views of real hyperspectral images and then uses these masks to supervise promptable hyperspectral segmentation. However, many images in Hyper-Seg are broad natural landscapes, such as mountains, rivers, and vegetation, where object contours are weak and semantic boundaries are often ambiguous. Moreover, the SAM-H-generated masks inevitably contain noisy or coarse regions, while HyperFree directly uses them for training without explicitly modeling such label uncertainty. This limits the boundary quality and object-centric supervision needed for a generalizable promptable \highlight{hyperspectral foundation model (HFM)}.

\textbf{Prior utilization bottleneck.} Many hyperspectral models use masked image modeling or reconstruction objectives to train a new backbone on hyperspectral data~\cite{hong2024spectralgpt,wang2024hypersigma}. This is reasonable when spectral statistics differ from those of RGB imagery, but it underuses the strong spatial priors already encoded in modern segmentation foundation models. SAM-style models learn to associate points, masks, boundaries, objectness, and multi-scale geometry through large visual corpora~\cite{kirillov2023sam,ravi2025sam2,carion2025sam3segmentconcepts}. These priors are highly relevant to hyperspectral remote sensing because many downstream tasks are not purely spectral. They require the model to propose coherent fields, rooftops, vehicles, roads, rivers, shorelines, changed parcels, or target-like regions. The key question is therefore not whether RGB priors should be reused, but how to inject spectral evidence without destroying them.

\begin{figure}[t]
    \centering
    \includegraphics[width=1\linewidth]{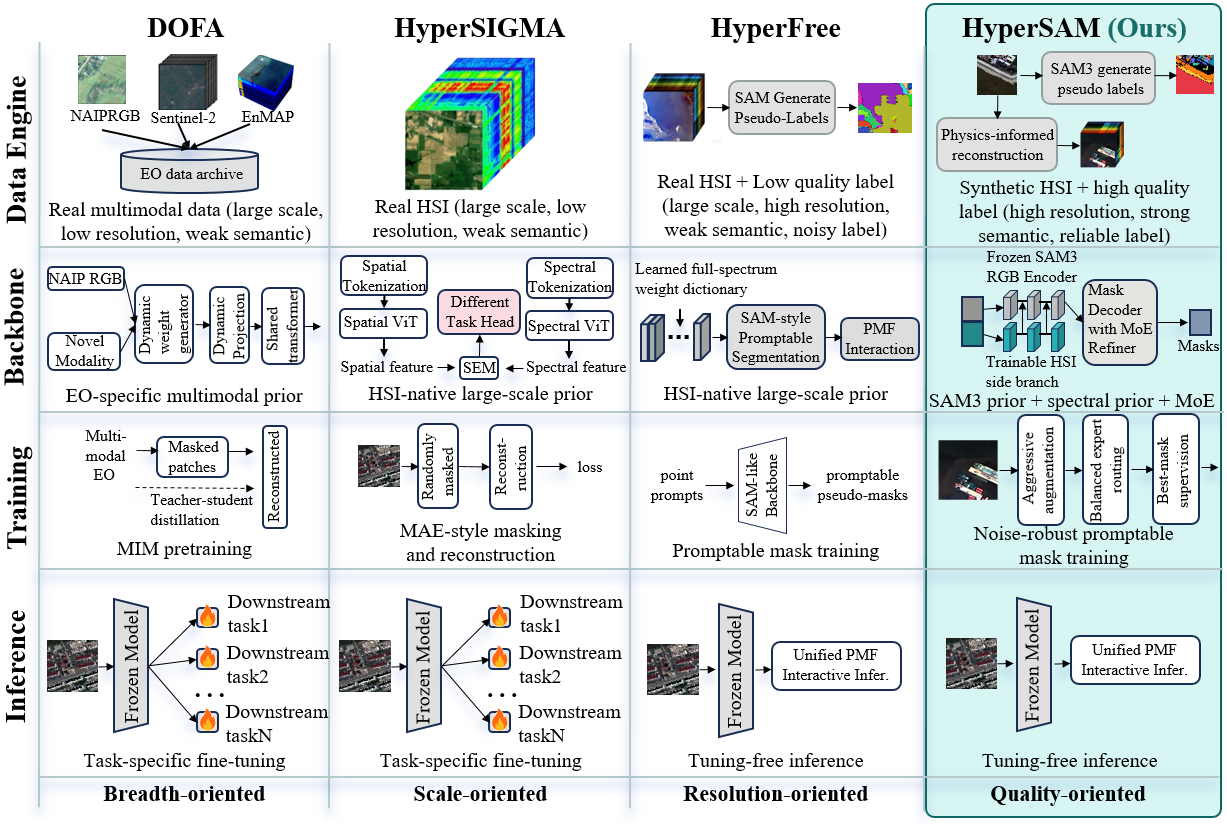}
    \caption{Comparison of data sources and training paradigms among representative hyperspectral foundation models. HyperSAM builds an object-centric hyperspectral training corpus by synthesizing full-spectrum HSI from high-resolution multispectral SpaceNet imagery and pairing it with SAM3-generated pseudo-masks.}
    \label{fig:comparison}
\end{figure}

To alleviate these issues, this paper proposes \textbf{HyperSAM}, a promptable hyperspectral foundation model that jointly addresses data quality, prior utilization, and noise-robust supervision. HyperSAM constructs an object-centric hyperspectral training corpus by synthesizing full-spectrum HSI from high-resolution multispectral SpaceNet imagery and pairing the reconstructed cubes with SAM3-derived pseudo-masks, thereby providing sharper spatial details and more reliable object-level supervision than existing hyperspectral foundation-model training corpora. It further adapts the SAM3 backbone to hyperspectral remote sensing through a prior-preserving dual-branch architecture, where a frozen RGB branch retains strong geometric and prompt-response priors while a trainable spectral branch injects full-spectrum corrections through zero-initialized residual adapters. A lightweight mixture-of-experts (MoE) decoder is introduced to refine task-agnostic mask predictions under diverse object scales and residual label uncertainty. To enhance robustness against imperfect pseudo-masks, HyperSAM further incorporates a noisy-label learning strategy inspired by Cross-modal Sample Selection (CromSS)~\cite{liu2025cromss}, which was originally proposed to mitigate noisy supervision in multimodal remote sensing segmentation by using cross-modal confidence masks to identify reliable training signals. In our setting, the RGB-prior branch and the hyperspectral branch provide complementary confidence cues for pseudo-mask supervision: high-confidence regions are emphasized, uncertain regions are softly down-weighted rather than treated as equally reliable, and class-balanced confidence selection is used to prevent dominant categories or easy regions from overwhelming the training process. This confidence-aware weighting is combined with hyperspectral-specific spatial and spectral augmentations, improving invariance to geometric perturbations, spectral response variations, and band-wise degradation. With these designs, HyperSAM provides a unified prompt-mask-feature interface for hyperspectral classification, anomaly detection, change detection, and target detection, achieving strong cross-task generalization as shown in Fig.~\ref{fig:radar}. Fig.~\ref{fig:comparison} summarizes the resulting difference between HyperSAM and representative hyperspectral foundation-model paradigms.

The contributions of this work are summarized as follows.

\begin{itemize}
\item We introduce a \textbf{data-centric paradigm} for hyperspectral foundation modeling by synthesizing full-spectrum hyperspectral data from semantically rich multispectral imagery through a physics-informed abundance-transfer mechanism. The resulting corpus provides sharp spatial details, diverse object semantics, and well-aligned spectral-mask supervision.

\item We develop \textbf{HyperSAM}, a promptable hyperspectral foundation model that adapts the SAM3 backbone through a dual-branch spectral feature injection scheme and a lightweight mixture-of-experts decoder. This design transfers strong generic visual priors into the hyperspectral domain while preserving sensitivity to full-spectrum material information.

\item We formulate pseudo-mask supervision as a \textbf{common robust-learning problem} in hyperspectral foundation modeling. A confidence-aware training strategy uses agreement across spectral views and branches to softly reweight unreliable regions \highlight{and improve} stability under noisy pseudo-labels.

\item We demonstrate the \textbf{cross-task \highlight{effectiveness}} of HyperSAM on multiple hyperspectral remote sensing tasks, including hyperspectral classification, anomaly detection, change detection, and target detection. \highlight{Under the common small-scene protocols, the experiments show strong transfer with a single frozen checkpoint and validate the} unified prompt-mask-feature inference paradigm.
\end{itemize}

\section{Related Work}
\label{sec:related}

\subsection{Remote Sensing and Hyperspectral Foundation Models}
Recent remote sensing foundation models have moved from task-specific learning toward reusable representations across sensors, scenes, and downstream tasks.
SpectralGPT explores spectral remote sensing pretraining with a 3D generative transformer and demonstrates transferability across scene classification, semantic segmentation, and change detection~\cite{hong2024spectralgpt}.
SkySense extends this trend to multimodal and multitemporal remote sensing by jointly modeling optical and synthetic aperture radar (SAR) observations with large-scale spatiotemporal pretraining~\cite{guo2024skysense}, while DOFA introduces a wavelength-conditioned dynamic framework to handle heterogeneous Earth observation sensors~\cite{xiong2024dofa}.
For hyperspectral imagery, HyperSIGMA builds a billion-scale HSI foundation model trained on HyperGlobal-450K and shows strong cross-task representational ability~\cite{wang2024hypersigma}.
SpectralEarth further highlights the importance of large-scale and globally distributed Environmental Mapping and Analysis Program (EnMAP) data for hyperspectral foundation-model training~\cite{braham2025spectralearth}.
\highlight{Generic self-supervised models such as DINOv3 follow a similar feature-centric downstream paradigm, although they are pretrained on different data and are not specifically designed for hyperspectral imagery~\cite{simeoni2025dinov3}. These methods primarily follow a representation-learning paradigm: their pretrained encoders are transferred through task-specific heads, linear probing, or downstream fine-tuning. Promptable models instead expose point-, box-, mask-, or concept-driven interfaces and can generate masks or transferable features without updating the backbone at inference time.} \highlight{More recently, HyperFree addresses channel-adaptive and tuning-free hyperspectral interpretation through prompt engineering and a learned wavelength dictionary~\cite{li2025hyperfree}.   It is therefore the closest prior promptable HSI method to HyperSAM.}

\highlight{The two promptable approaches nevertheless differ in data, supervision, and spectral adaptation. HyperFree is trained on real Hyper-Seg images using SAM-H-derived pseudo-masks as direct supervision. HyperSAM constructs physically constrained full-spectrum synthetic cubes from high-spatial-resolution SpaceNet imagery, preserves a frozen SAM3 RGB branch, injects material-sensitive features through a trainable hyperspectral side encoder with zero-initialized adapters, refines masks through a scale-adaptive MoE path, and treats SAM3-derived pseudo-masks as noisy supervision through confidence-aware reweighting. HyperSAM therefore targets} object-centric promptable \highlight{learning} with high spatial fidelity \highlight{while retaining the original prompt-response prior}.

\subsection{Promptable Segmentation for Remote Sensing}
Promptable segmentation models provide a flexible interface for generating masks from points, boxes, masks, or semantic prompts.
SAM introduced general promptable segmentation, while SAM3 further extends the paradigm to concept-level prompting by detecting, segmenting, and tracking all instances matching a given concept~\cite{kirillov2023sam,carion2025sam3segmentconcepts}.
In remote sensing, RSPrompter learns prompts for SAM-based instance segmentation~\cite{chen2024rsprompter}, PointSAM adapts SAM with point-level supervision and pseudo-label refinement~\cite{liu2025pointsam}, and AnyChange exploits SAM latent matching for zero-shot remote sensing change detection~\cite{zheng2024anychange}.
RemoteSAM further formulates Earth observation understanding as a referring-segmentation-centered framework for unifying pixel-, region-, and image-level tasks~\cite{yao2025remotesam}.
These methods demonstrate the value of promptable segmentation priors in remote sensing, but they are mainly designed for RGB or multispectral imagery.
HyperSAM differs by preserving the geometric and prompt-response priors of SAM3 while injecting full-spectrum hyperspectral information through a trainable spectral branch.

\subsection{Synthetic Hyperspectral Data and Noisy-Label Learning}
Large-scale dense annotation remains a major obstacle for hyperspectral foundation-model training. Physics-informed reconstruction methods synthesize hyperspectral data from multispectral observations by estimating material abundances and transferring them through spectral endmember libraries~\cite{bioucas2012hyperspectral,liu2022pdass,kokaly2017usgsspectral}. HyperSAM follows this physically constrained direction, but uses it for foundation-model data construction: pseudo-masks are obtained from high-resolution visible imagery, while full-spectrum cubes are reconstructed from aligned multispectral observations. Because generated masks remain imperfect, robust learning from noisy pseudo-labels is also required. Co-teaching, confident learning, and broad noisy-label studies show that treating all labels as equally reliable can cause confirmation bias and degraded generalization~\cite{han2018coteaching,northcutt2021confident,song2023noisylabels}. In multimodal remote sensing, Cross-modal Sample Selection uses cross-modal confidence and entropy cues to select reliable noisy supervision~\cite{liu2025cromss}. HyperSAM adapts this general principle to promptable hyperspectral training by using spectral-view and branch-level agreement to construct soft, class-balanced confidence weights.

\section{Method Overview}
\label{sec:method}

HyperSAM is built around three coupled components: high-quality hyperspectral dataset construction, a SAM3-based promptable model framework, and a robust learning strategy for noisy pseudo-mask supervision.
The dataset construction component synthesizes full-spectrum hyperspectral cubes from high-resolution multispectral SpaceNet imagery through a physics-informed abundance-transfer mechanism and pairs them with SAM3-derived object masks from the corresponding RGB images.
The model framework adapts SAM3 to hyperspectral imagery by preserving the frozen RGB prior branch while using a trainable spectral branch to inject full-spectrum information through zero-initialized residual adapters~\cite{zhang2023adding}. A lightweight MoE decoder further refines mask predictions across different object scales and scene structures~\cite{shazeer2017moe,fedus2022switch}.
The robust learning component handles imperfect pseudo-masks through hyperspectral-specific augmentations and confidence-aware noisy-label weighting inspired by Cross-modal Sample Selection (CromSS)~\cite{liu2025cromss}, emphasizing reliable regions while softly down-weighting uncertain ones.
Finally, HyperSAM predicts task-agnostic masks, mask quality scores, and transferable feature representations that support downstream hyperspectral interpretation. Details are provided in the following subsections.

\subsection{High-Fidelity Hyperspectral Dataset Construction}
\label{sec:data}
\begin{figure}[t]
\centering
\includegraphics[width=1\linewidth]{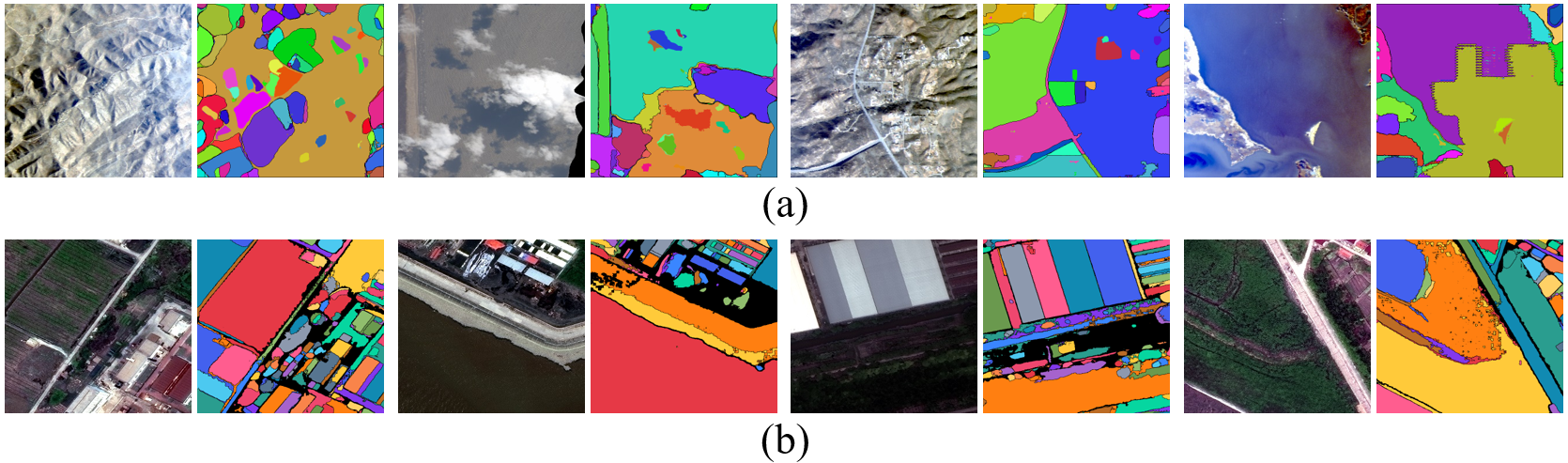}
\caption{Visual comparison of training data. HyperFree samples are often limited by coarse spatial details or weaker object boundaries, whereas the HyperSAM synthetic training data contain sharper outlines, richer object categories, and pseudo-masks suitable for promptable foundation-model training.}
\label{fig:dataset_comp}
\end{figure}

The first component constructs an object-centric hyperspectral training corpus from high-resolution multispectral imagery.
We use SpaceNet~\highlight{2} imagery~\cite{vanetten2019spacenetremotesensingdataset}, which provides aligned RGB images and 8-band WorldView-3 multispectral observations over \highlight{four geographically distinct urban regions: Las Vegas, Paris, Shanghai, and Khartoum. One synthetic hyperspectral cube and its corresponding pseudo-label set are generated from each retained aligned RGB--multispectral patch. The resulting corpus contains 10,592 training patches and 3,527 validation patches, providing variation in urban morphology, surface materials, vegetation, and environmental conditions}.
This choice is motivated by two complementary advantages.
First, the RGB images contain sharp spatial structures and rich urban objects, allowing SAM3 to generate more reliable object-level pseudo-masks than those obtained from low-resolution or weakly structured hyperspectral scenes.
Second, the multispectral observations preserve physically meaningful spectral responses of the same surface materials, making them suitable for constrained hyperspectral reconstruction rather than purely appearance-based synthesis.
As shown in Fig.~\ref{fig:dataset_comp}, the resulting training samples contain clearer object boundaries and richer semantic entities than existing pseudo-labeled hyperspectral corpora.

In our construction, semantic annotation and spectral reconstruction are decoupled by assigning distinct roles to different spectral bands of the input data. Object masks are extracted from the visible (RGB) bands of the high-resolution multispectral images, where promptable segmentation models perform reliably. Concurrently, the full sets of multispectral channels are converted into hyperspectral cubes through a physics-informed abundance-transfer process. This design avoids applying segmentation models directly to non-visible spectral subsets with ambiguous object boundaries, effectively preserving high-quality spatial supervision while enriching each pixel with full-spectrum material information.

\begin{figure}[t]
\centering
\includegraphics[width=1\linewidth]{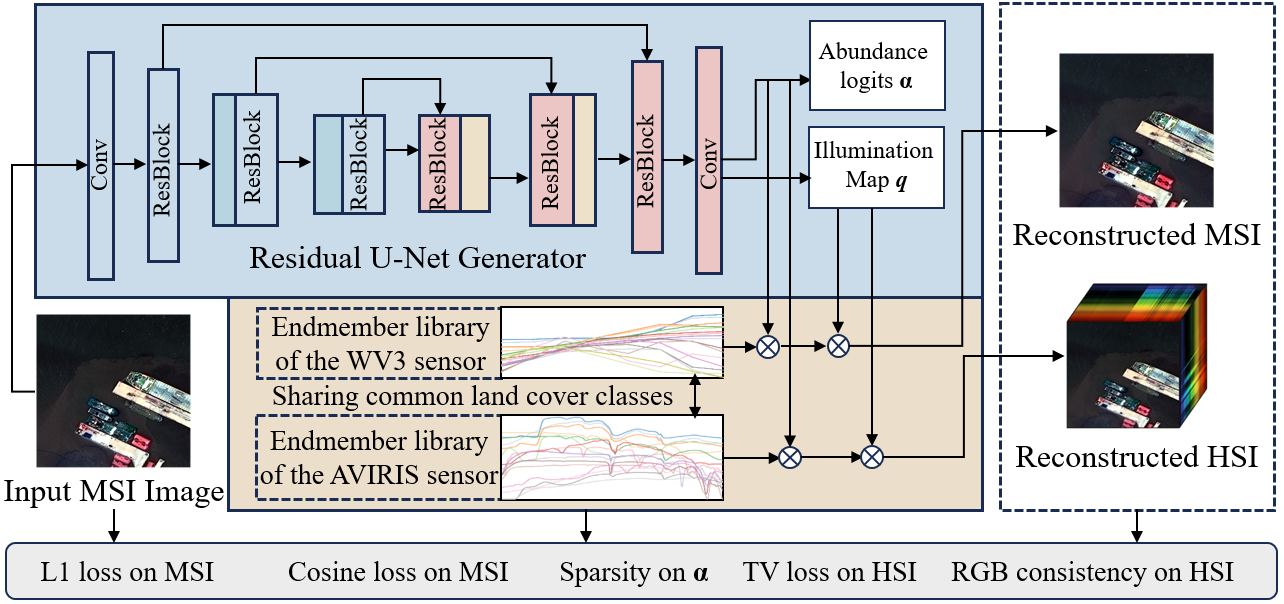}
\caption{Physics-informed hyperspectral reconstruction pipeline. Multispectral SpaceNet imagery is decomposed into abundance and illumination maps, which are transferred through a hyperspectral endmember library to synthesize full-spectrum cubes aligned with object-level pseudo-masks.}
\label{fig:reconstruction}
\end{figure}

The reconstruction module follows the physics-informed hyperspectral synthesis principle of Physics-informed Deep Adversarial Spectral Synthesis (PDASS)~\cite{liu2022pdass}. The overall reconstruction process is illustrated in Fig.~\ref{fig:reconstruction}.
Rather than regressing hyperspectral bands independently, we assume that both multispectral and hyperspectral observations of the same scene can be explained by shared material abundances under different sensor response functions.
This assumption is consistent with the linear spectral mixing model widely used in hyperspectral analysis: a pixel spectrum is represented as a mixture of several material endmembers modulated by illumination.
Therefore, once the material abundance and illumination of each pixel are estimated from the multispectral image, the same physical composition can be transferred to a hyperspectral endmember library to synthesize a full-spectrum cube.
The source endmember spectra are \highlight{obtained} from the public \textit{USGS Spectral Library Version 7} data release~\cite{kokaly2017usgsspectral}. \highlight{Following the physics-informed selection strategy of PDASS~\cite{liu2022pdass}, we retain artificial materials, coatings, liquids, minerals, organic compounds, soils and mixtures, and vegetation. A spectrum is retained only when a matched, complete record exists in both the WorldView-3-resampled and Airborne Visible/Infrared Imaging Spectrometer (AVIRIS)-2014-convolved releases and every reflectance value lies in $[0,1.5]$. Uniform subsampling by a factor of two yields $E=772$ paired materials. We use these sensor-specific releases directly and match their records by material identifier, producing} $A_{\ms}\in\R^{8\times E}$ and $A_{\hsi}\in\R^{224\times E}$ \highlight{whose corresponding columns describe the same material. The AVIRIS library covers approximately 400--2500~nm}.

Specifically, given an 8-band multispectral patch $Y\in\R^{8\times H\times W}$, a U-Net-style conditional generator predicts abundance logits and an illumination term:
\begin{equation}
    Z=G_{\phi}(Y),\quad Z\in\R^{(E+1)\times H\times W}.
\label{eq:generator}
\end{equation}
For each pixel $p$, the abundance vector is constrained to lie on a simplex and the illumination is constrained to be positive:
\begin{equation}
    a_e(p)=\frac{\exp Z_e(p)}{\sum_{e'=1}^{E}\exp Z_{e'}(p)},
    \quad
    \ell(p)=\mathrm{softplus}(Z_{E+1}(p))+\epsilon.
\label{eq:abundance}
\end{equation}
The multispectral reconstruction and hyperspectral synthesis are then obtained by
\begin{equation}
    \hat{Y}(:,p)=\ell(p)A_{\ms}a(p),\quad
    X^{\syn}(:,p)=\ell(p)A_{\hsi}a(p).
\label{eq:hsi_recon}
\end{equation}
In this way, the synthesized hyperspectral cube is generated through material-consistent abundance transfer rather than unconstrained image-to-image translation.
The multispectral branch ensures that the reconstructed spectrum remains faithful to the observed WorldView-3 measurements, while the hyperspectral branch expands the same material composition to dense spectral bands.

The reconstruction generator is optimized before HyperSAM training and then kept fixed when constructing the corpus.
Its objective combines multispectral fidelity, spectral-shape consistency, projection consistency, abundance sparsity, and spatial smoothness:
\begin{equation}
\begin{aligned}
    \mathcal{L}_{\mathrm{rec}}
    =&\ \lambda_{1}\|\hat{Y}-Y\|_{1}
    +\lambda_{\cos}\mathcal{L}_{\cos}(\hat{Y},Y)
    +\lambda_{\mathrm{proj}}\|P_{\ms}X^{\syn}-Y\|_{1}  \\
    &+\lambda_{\mathrm{sp}}\mathcal{L}_{1/2}(a)
    +\lambda_{\mathrm{tv}}\left(\mathrm{TV}(a)+\mathrm{TV}(\ell)\right),
\end{aligned}
\label{eq:recon_loss}
\end{equation}

where $Y$ and $\hat{Y}$ denote the observed and reconstructed multispectral images, respectively, and $X^{\text{syn}}$ represents the synthesized hyperspectral cube. $P_{\text{ms}}$ denotes the multispectral sensor-response projection from $X^{\text{syn}}$ to the WorldView-3 bands. The variables $a$ and $\ell$ represent the abundance map and the illumination map, while $\mathcal{L}_{1/2}(\cdot)$ and $\text{TV}(\cdot)$ denote the $L_{1/2}$ abundance-sparsity regularizer and the total-variation regularization, respectively. The loss weights are fixed as $(\lambda_1,\lambda_{\cos},\lambda_{\mathrm{proj}},\lambda_{\mathrm{sp}},\lambda_{\mathrm{tv}})=(100,1000,10,30,10)$ for all corpus-generation runs and are not tuned on any downstream benchmark. The $L_1$ term preserves observed WorldView-3 radiometry, the cosine term preserves spectral shape, the projection term requires the synthesized spectrum to reproduce the measured multispectral observation through the sensor-response operator, the $L_{1/2}$ abundance-sparsity term encourages parsimonious material mixtures, and total variation regularizes the abundance and illumination maps spatially. These complementary constraints make the synthesis physically interpretable and reduce the risk of generating spectrally implausible training samples.

 \highlight{Both the pseudo-mask generator and the frozen RGB-prior branch use Meta's original November 2025 SAM3 image release~\cite{carion2025sam3segmentconcepts}, implemented with the official codebase}\footnote{\url{https://github.com/facebookresearch/sam3}} \highlight{and the released} \texttt{\highlight{facebook/sam3}} \texttt{\highlight{sam3.pt}} \highlight{checkpoint}\footnote{\url{https://huggingface.co/facebook/sam3}}\highlight{. The same checkpoint is used throughout without additional natural-image, remote-sensing, or hyperspectral pretraining. During HyperSAM training, one point prompt is sampled for each target instance and iterative prompt refinement is used to progressively correct the predicted mask.}

 Finally, each synthesized hyperspectral cube is paired with the SAM3-derived pseudo-masks from its aligned RGB image:
\begin{equation}
    \mathcal{D}=\{(X_i^{\syn},\mathcal{M}_i)\}_{i=1}^{N},\quad
    \mathcal{M}_i=\{M_{i1}^{*},\ldots,M_{iK_i}^{*}\}.
\label{eq:dataset}
\end{equation}
The resulting corpus combines the spatial clarity of high-resolution RGB segmentation with physically constrained hyperspectral signatures. This provides HyperSAM with training pairs that are object-centric, spectrally informative, and better aligned with promptable mask learning than conventional hyperspectral datasets.

\begin{figure}[t]
  \centering
  \includegraphics[width=\linewidth]{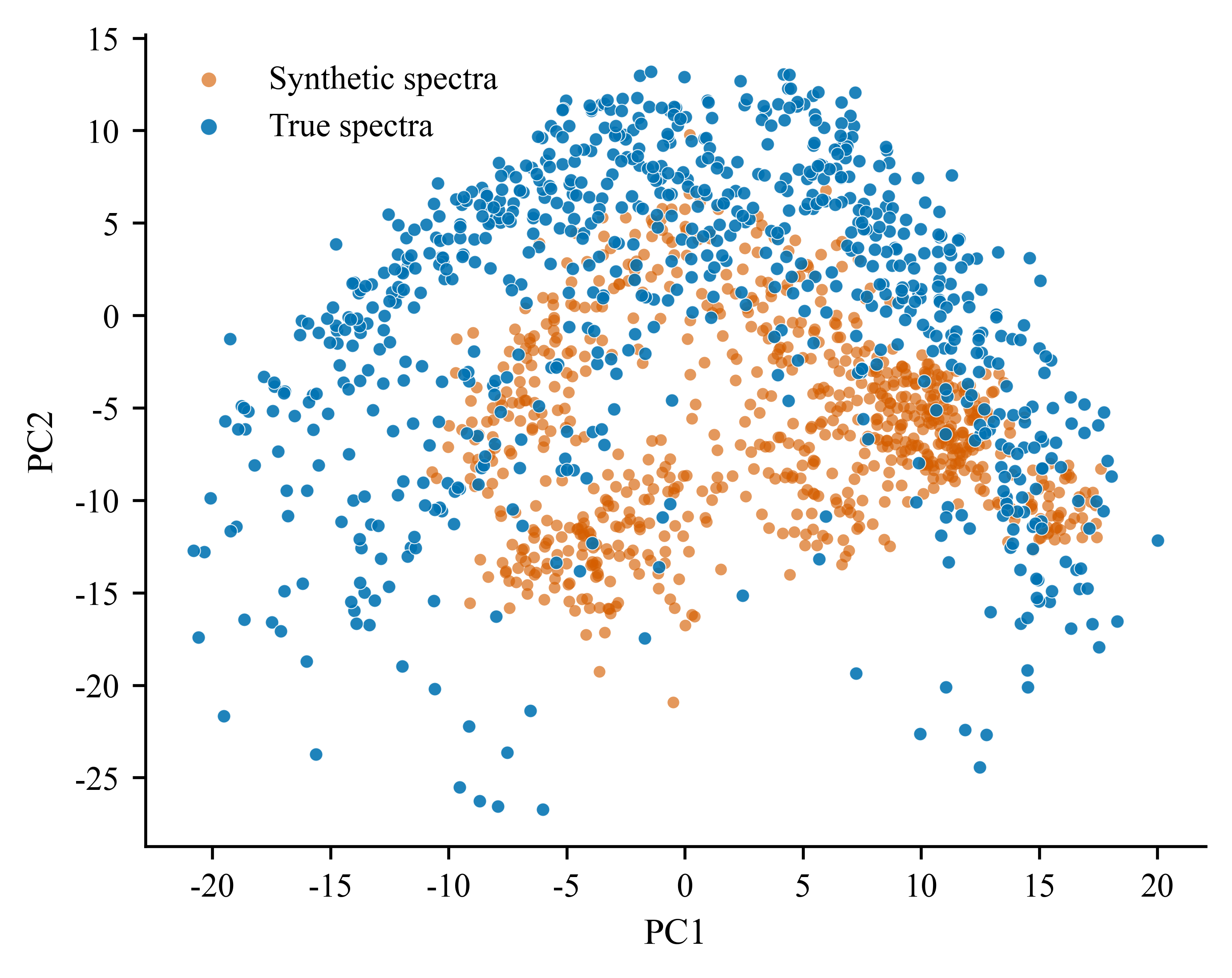}
  \caption{\highlight{Shared-PCA visualization of 800 randomly sampled synthetic spectra and 800 real AVIRIS spectra from Hyper-Seg. The core overlap indicates physically plausible synthetic coverage.}}
  \label{fig:synthesis_pca}
\end{figure}

\begin{figure*}[t]
  \centering
  \includegraphics[width=0.95\textwidth]{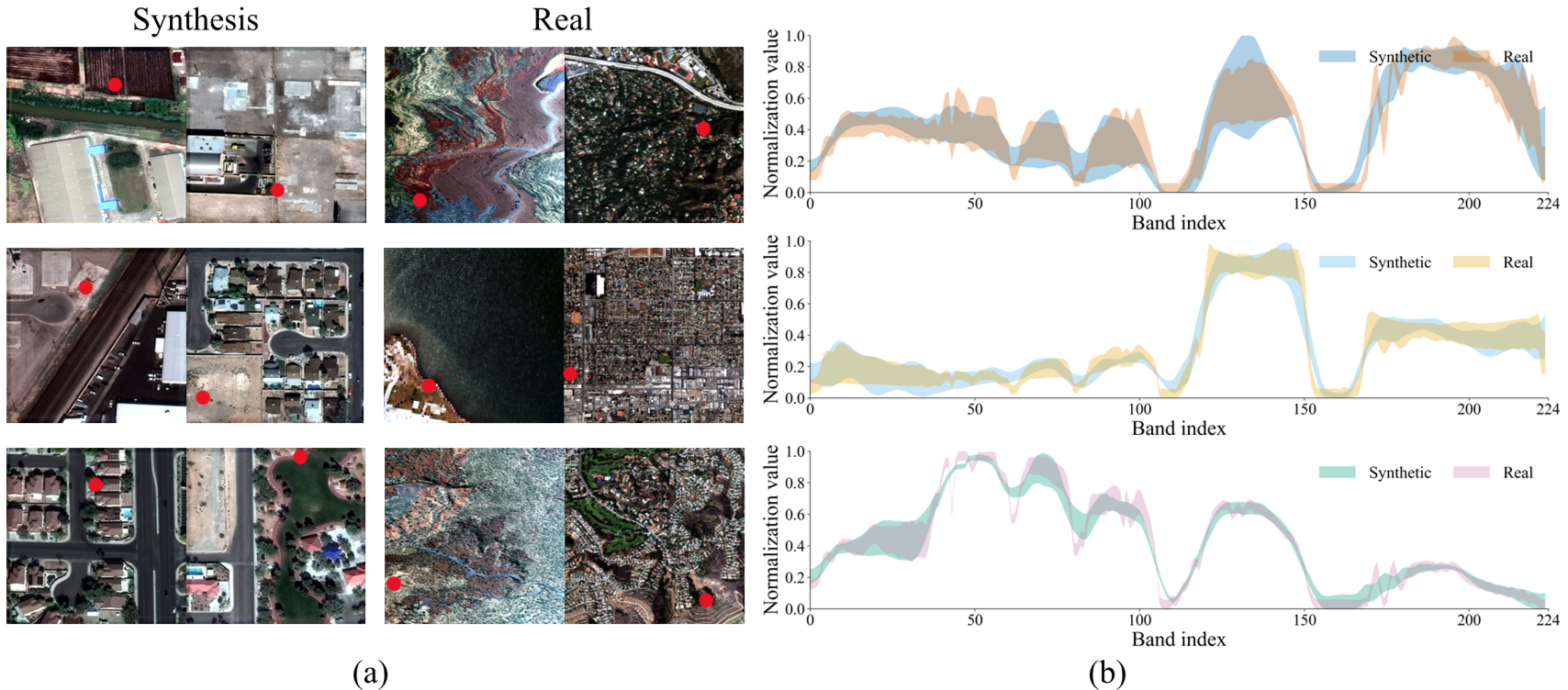}
  \caption{\highlight{Spatially grounded comparison between synthesized and real hyperspectral data. Left: representative synthetic and real AVIRIS image examples, with red dots marking sampled locations. Right: 25th--75th percentile spectral envelopes formed after independently normalizing each spectrum. The overlapping shapes provide qualitative evidence of material-dependent consistency while retaining visible distributional differences.}}
  \label{fig:synthesis_match}
\end{figure*}

 \highlight{To assess the synthetic-to-real spectral relationship, we randomly sample 800 synthesized spectra and 800 real AVIRIS spectra from Hyper-Seg~\cite{li2025hyperfree}
and project both sets with a shared PCA basis. Fig.~\ref{fig:synthesis_pca} shows substantial overlap in the core manifold, indicating that the synthetic samples occupy physically plausible regions. Fig.~\ref{fig:synthesis_match} compares independently normalized spectral envelopes for representative land-cover materials. Similar overall shapes and overlapping interquartile ranges provide complementary qualitative evidence of material-dependent consistency.}

\subsection{HyperSAM Model Architecture}
\label{sec:network}

HyperSAM adapts SAM3 to hyperspectral imagery by preserving its promptable segmentation priors while adding a lightweight spectral adaptation pathway. \highlight{HyperSAM retains the pretrained SAM3 image encoder, prompt encoder, and mask decoder with frozen parameters. Hyperspectral adaptation is achieved through a spectral-spatial patch embedding and spectral-shape extractor, a trainable side transformer, layer-wise zero-initialized residual adapters, and an MoE-based residual refiner.}

\begin{figure*}[t]
    \centering
    \includegraphics[width=0.95\linewidth]{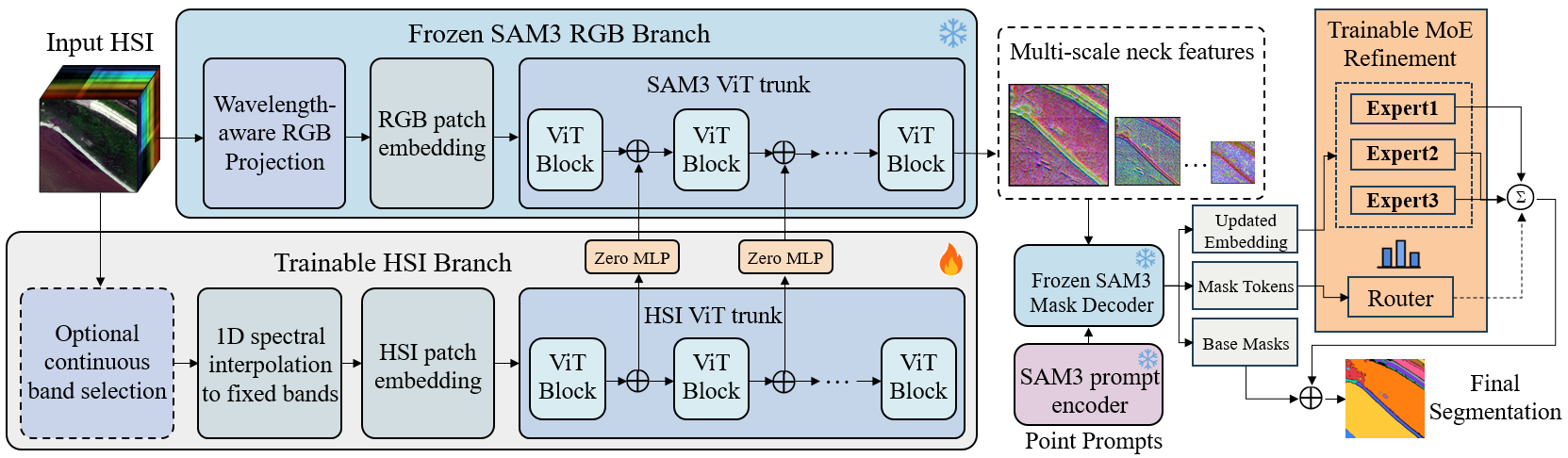}
    \caption{HyperSAM architecture. A frozen SAM3 RGB branch preserves objectness, boundary, and prompt-response priors, while a trainable hyperspectral side encoder injects full-spectrum residual features through zero-initialized adapters. The mask decoder is augmented with a lightweight multi-scale MoE refinement path for adapting mask logits and quality scores to hyperspectral imagery.}
    \label{fig:hypersam}
\end{figure*}

 As shown in Fig.~\ref{fig:hypersam}, the model contains four main parts: a frozen RGB prior branch, a trainable hyperspectral side branch, residual spectral injection adapters, and a prompt-driven mask decoder with MoE refinement. The RGB branch takes a three-channel proxy \highlight{formed from the closest available bands to 700, 546.1, and 438.8~nm. Nearest-band selection, percentile stretching, the official SAM3 normalization, and resizing to $1008\times1008$ are applied before the proxy is passed} through the frozen SAM3 image encoder\highlight{. Feature level~2 is used for mask--feature matching}. In parallel, the hyperspectral cube is processed by the side branch to extract full-spectrum material information that is unavailable in the RGB proxy. \highlight{During training, all SAM3 image-encoder and prompt-processing parameters remain frozen, while the hyperspectral side encoder, injection adapters, and MoE refinement path are optimized.}

For the hyperspectral branch, inputs from different sensors are first interpolated to a fixed spectral dimension.
A spectral-spatial patch embedding mixes local spatial context with band-wise information, while a spectral-shape extractor models the reflectance profile at each spatial location.
The resulting tokens are projected to the same dimension and spatial resolution as the SAM3 visual tokens, so that spectral evidence can be injected into the frozen RGB representation layer by layer.
The side encoder is initialized from the RGB transformer where dimensions are compatible, which provides a stable starting point for spectral adaptation.

Let $R_l$ denote the frozen RGB feature at layer $l$, $S_l$ denote the hyperspectral side feature, $B_l(\cdot)$ denote the frozen SAM3 transformer block, $B_l^{\hsi}(\cdot)$ denote the side-branch transformer block, and $Z_l(\cdot)$ denote a zero-initialized residual adapter.
The spectral injection process is written as
\begin{align}
    S_0 &= R_0 + W_0\,\mathrm{Patch}_{\hsi}(X), \\
    S_{l+1} &= B_l^{\hsi}(S_l), \\
    R_{l+1} &= B_l(R_l)
    + \mathbf{1}[l\ge l_0] Z_l(S_{l+1}).
\label{eq:inject}
\end{align}
Here $W_0$ and $Z_l$ are initialized to produce zero residuals.
Therefore, the adapted model is initially equivalent to the original SAM3 RGB branch, and spectral corrections are gradually learned only when they improve hyperspectral mask prediction.
This identity-preserving design reduces the risk of destroying the inherited SAM3 priors during adaptation.

The prompt encoder remains compatible with the SAM3 interface and supports sparse prompts such as points and boxes, as well as dense mask prompts.
The mask decoder receives the fused image features and prompt embeddings and produces base mask logits $M^0$, mask quality scores $q^0$, mask tokens $T$, and an upscaled feature map $U$.
To refine masks under different object scales and scene layouts, HyperSAM integrates a lightweight MoE refiner inside the decoder.
A soft router predicts expert weights from the mean-pooled mask tokens:
\begin{equation}
    \pi=\mathrm{softmax}
    \left(
    g\left(\frac{1}{K_m}\sum_{j=1}^{K_m}T_j\right)
    \right),
\label{eq:router}
\end{equation}
where $K_m$ is the number of mask tokens. \highlight{We use $K_e=3$ convolutional} experts with different receptive fields \highlight{to} predict residual mask logits:
\begin{equation}
\begin{aligned}
    \Delta M &= \sum_{k=1}^{K_e}\pi_k E_k(U),\\
    M &= M^0 + \Delta M.
\end{aligned}
\label{eq:moe_delta}
\end{equation}
A zero-initialized residual head also adjusts the mask quality score.
The MoE module is used for scale-adaptive mask refinement, not for noisy-label correction.
To prevent the router from collapsing to a single expert, we apply an expert-balance loss:
\begin{equation}
    \mathcal{L}_{\mathrm{bal}}
    = K_e \sum_{k=1}^{K_e}
    \left(
    \frac{1}{B}\sum_{b=1}^{B}\pi_{bk}
    \right)^2.
\label{eq:balance}
\end{equation}

After training, given a hyperspectral image and a prompt, HyperSAM produces three unified outputs: task-agnostic masks, mask quality scores, and dense spectral-spatial feature embeddings.
Specifically, the masks describe prompt-related candidate regions, the quality scores rank their reliability, and the dense embeddings provide material- and context-aware features for similarity matching, prototype construction, and temporal comparison.
The same prompt-mask-feature interface is then reused for hyperspectral classification, anomaly detection, change detection, and target detection.

\subsection{Robust Learning With Confidence-Aware Noisy-Label Weighting}
\label{sec:training}

Although SAM3-derived masks provide stronger object boundaries than coarse hyperspectral pseudo-labels, they may still contain uncertain boundaries, incomplete objects, fragmented regions, or background leakage.
 \highlight{The RGB-derived masks are nevertheless treated as potentially noisy supervision rather than exact hyperspectral ground truth: visible object boundaries and material boundaries can disagree around shadows, thin structures, mixed pixels, roof--background transitions, and visually similar surfaces with different spectra. Large homogeneous object interiors are generally more reliable because the RGB image and synthesized cube share the same geometry, while confidence-aware training locally suppresses ambiguous boundaries and background leakage rather than rejecting an entire mask.}
 HyperSAM therefore adopts a robust learning strategy that combines hyperspectral-specific augmentations, best-mask supervision, cross-view consistency, and confidence-aware noisy-label weighting.

Specifically, Cross-modal Sample Selection (CromSS)~\cite{liu2025cromss} selects reliable noisy supervision by comparing confidence estimates from different modalities and reducing the influence of uncertain regions.
HyperSAM extends this idea from multimodal segmentation to promptable hyperspectral learning.
Instead of relying on external modalities, it constructs two augmented spectral views from the same hyperspectral cube and uses the RGB-prior branch and hyperspectral branch as complementary confidence sources.
Regions that remain confident and consistent across views and branches receive larger weights, while ambiguous regions are softly down-weighted rather than discarded.
Class-balanced selection is further used to prevent frequent background regions or easy objects from dominating the supervision.

For each pseudo-mask $M^*$, the decoder predicts multiple candidate masks $\{M_j\}$.
HyperSAM follows SAM-style best-mask supervision and selects the candidate with the smallest confidence-weighted sum of binary cross-entropy (BCE) and Dice losses:
\begin{align}
    \mathcal{L}_{j}^{\mathrm{seg}}
    &= \mathcal{L}_{\mathrm{BCE}}(M_j,M^*;W)
    + \mathcal{L}_{\mathrm{Dice}}(M_j,M^*;W),\\
    j^* &= \arg\min_j \mathcal{L}_{j}^{\mathrm{seg}}.
\label{eq:bestmask}
\end{align}
The weighted Dice loss is defined as
\begin{equation}
\mathcal{L}_{\mathrm{Dice}}
=1-
\frac{
2\langle W\odot\sigma(M_j),M^*\rangle+1
}{
\langle W,\sigma(M_j)\rangle
+\langle W,M^*\rangle+1
}.
\label{eq:weighted_dice}
\end{equation}
Here $W$ is the confidence weight map.
Given a binary pseudo-label $y(p)\in\{0,1\}$ and the corresponding mask logit $z(p)$, the label-class confidence is
\begin{equation}
f(p)=
\begin{cases}
\sigma(z(p)), & y(p)=1,\\
1-\sigma(z(p)), & y(p)=0.
\end{cases}
\label{eq:label_conf}
\end{equation}
Foreground and background pixels are processed separately.
Within each class, pixels above the class-specific confidence threshold receive full weight, while the remaining pixels are assigned smaller weights according to their confidence.
The selected proportion is controlled by a training schedule that gradually shifts from full supervision to confidence-aware supervision.

To build cross-view confidence, HyperSAM samples two \highlight{contiguous} spectral windows from the same synthetic cube\highlight{. Each window length is drawn uniformly from 128 to 224 bands, and its starting position is sampled uniformly from the valid range. The selected window is then interpolated to 224 channels for the common model interface. The second view is sampled independently, redrawn if it is identical to the first, and constrained to share at least 32 bands with the first view. The minimum overlap preserves common material evidence, while the wide random windows and independent starts introduce substantial spectral-response variation. Both views retain the same spatial target, so the consistency term regularizes the model against spectral cropping rather than changing the segmentation supervision}.
Let $f_a$ and $f_b$ be the label-class confidence maps from the two views.
A CromSS-style common-confidence update is applied as
\begin{equation}
\begin{aligned}
    \hat{f}_a &= \frac{1}{2}(f_a+f_a f_b),\\
    \hat{f}_b &= \frac{1}{2}(f_b+f_a f_b).
\end{aligned}
\label{eq:common_conf}
\end{equation}
The enhanced confidence maps are used to construct the weight maps for the two supervised losses.
In parallel, a symmetric Kullback--Leibler (KL) divergence term encourages the two spectral views to produce consistent binary mask distributions:
\begin{equation}
\mathcal{L}_{\mathrm{cons}}
=
\frac{1}{2}
\left[
D_{\mathrm{KL}}(p_a\|p_b)
+
D_{\mathrm{KL}}(p_b\|p_a)
\right],
\label{eq:consistency}
\end{equation}
where
\begin{equation}
    p_a=[1-\sigma(M_a),\sigma(M_a)],
    \quad
    p_b=[1-\sigma(M_b),\sigma(M_b)].
\label{eq:binary_dist}
\end{equation}
Entropy-derived confidence weights are used so that stable regions contribute more strongly to the consistency constraint.

\begin{figure}[t]
\centering
\includegraphics[width=\linewidth]{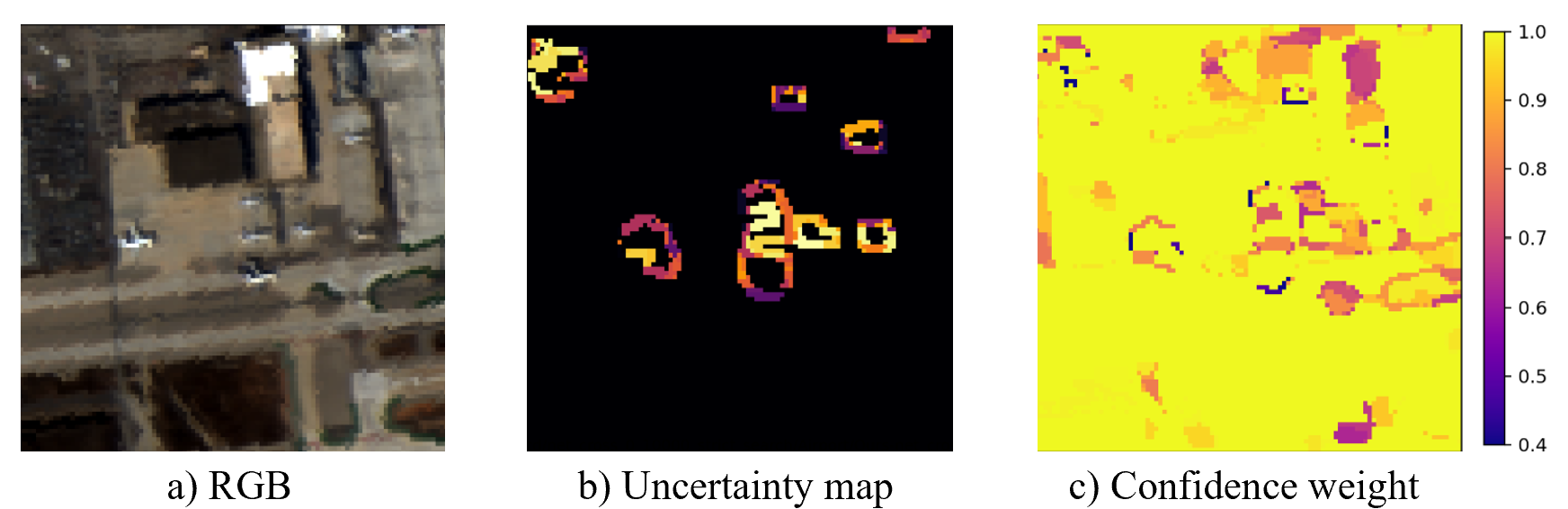}
\caption{ \highlight{Representative analysis} of \highlight{pseudo-mask uncertainty and} confidence-aware \highlight{weighting.} (\highlight{a}) \highlight{RGB image used to provide} the \highlight{SAM3 spatial prior. (b) Estimated} uncertainty map, \highlight{in which unreliable regions concentrate around} ambiguous boundaries\highlight{, fragmented objects,} and \highlight{complex} background \highlight{structures. (c) Confidence-weight map}, \highlight{which assigns larger weights to stable regions and smaller weights to uncertain pseudo-label pixels}.}
\label{fig:weights_vis}
\end{figure}

 \highlight{For each view, the selected best-mask loss is the equally weighted sum of confidence-weighted sigmoid BCE and Dice losses in Eq.~\eqref{eq:bestmask}. The two selected view losses are averaged as}
\begin{equation}
\highlight{\mathcal{L}_{\mathrm{seg}}
=\frac{1}{2}\left(
\mathcal{L}_{\mathrm{mask}}^{a}
+\mathcal{L}_{\mathrm{mask}}^{b}
\right).
\label{eq:seg_loss}}
\end{equation}
 The final training objective is
\begin{equation}
 \highlight{\mathcal{L}
=\mathcal{L}_{\mathrm{seg}}
+\lambda_{\mathrm{cons}}\mathcal{L}_{\mathrm{cons}}
+\lambda_{\mathrm{bal}}\mathcal{L}_{\mathrm{bal}},}
 \label{eq:full_loss}
\end{equation}
 \highlight{where $\lambda_{\mathrm{cons}}=0.1$ and $\lambda_{\mathrm{bal}}=0.05$ in all experiments. Here $\mathcal{L}_{\mathrm{cons}}$ improves invariance across spectral windows, while} $\mathcal{L}_{\mathrm{bal}}$ \highlight{prevents expert collapse}. Together with spatial rotations, sparse spatial masking, random spectral-window sampling, and label smoothing, this robust learning strategy reduces the influence of unreliable pseudo-mask regions while preserving useful supervision from high-confidence regions. As illustrated in Fig.~\ref{fig:weights_vis}, the resulting confidence-aware weight map assigns large weights to reliable object regions, while ambiguous boundaries and noisy background fragments are highlighted as uncertain areas.

\section{Experiments}
\label{sec:experiments}

We evaluate HyperSAM on four representative hyperspectral interpretation tasks: hyperspectral classification (HC), hyperspectral anomaly detection (HAD), hyperspectral change detection (HCD), and hyperspectral target detection (HTD). The main tables test whether the promptable spectral adaptation improves cross-task generalization under a single checkpoint, while the ablation studies isolate the data synthesis pipeline, promptable spectral branch, MoE refinement, and robust pseudo-mask weighting. Following the task-oriented organization commonly used in hyperspectral foundation-model evaluation~\cite{wang2024hypersigma}, each task first specifies the dataset, supervision, prompting rule, and decision rule, and then reports quantitative and qualitative results. All experiments use the same trained HyperSAM checkpoint. No task-specific network is trained for any downstream benchmark. \highlight{The evaluation is intentionally limited to one-shot, zero-shot, or otherwise low-label small-scene settings and does not constitute a large-patch benchmark with full downstream fine-tuning.}

In addition to reporting final accuracy, the experimental analysis is organized to echo the two bottlenecks identified earlier: whether high-quality spectral-mask data are more useful than merely scaling weak supervision, and whether promptable visual priors can be reused without losing hyperspectral material sensitivity. Therefore, the following analysis discusses not only which method obtains the best metric, but also why the same prompt-mask-feature interface helps or fails under different task assumptions. The ablations further separate the contributions of synthetic spectral construction, MoE mask refinement, and confidence-aware pseudo-mask weighting, so that the practical role of each component can be interpreted beyond a simple state-of-the-art comparison.

\subsection{General Experimental Settings}

\subsubsection{Datasets and Metrics}
We report compact comparisons on two representative datasets for each task. HC is evaluated on Indian Pines~\cite{baumgardner2015indianpines} and Pavia University~\cite{gic2021paviau}, which contain 16 agricultural land-cover classes and 9 urban land-cover classes, respectively. HAD is evaluated on Beach-1 and Beach-2 from the Airport-Beach-Urban benchmark~\cite{tu2019sdbp}. HCD is evaluated on BayArea/River and Hermiston following standard hyperspectral change-detection protocols~\cite{ml_edan_2022,wang2022sstformer}. HTD is evaluated on the Nuance Cri scene~\cite{zhang2017ajsrm} and the Airport scene from the Airport-Beach-Urban benchmark~\cite{tu2019sdbp}. For HC, we report overall accuracy (OA), average accuracy (AA), and Kappa. For HAD and HTD, we report detection factor (DF) and overall detection performance (ODP). For HCD, we report intersection-over-union (IoU) and F1 score. All reported metrics follow the higher-is-better convention.

We additionally evaluate HyperSAM on the Hyperspectral Oil Spill Database (HOSD) benchmark for hyperspectral oil-spill mapping~\cite{duan2023hosd}. Specifically, we use the Gulf of Mexico GM13 and GM17 scenes, where AVIRIS hyperspectral cubes provide dense spectral measurements for distinguishing oil-covered water from clean or non-oil sea surfaces. Each pixel is annotated as either oil spill or clean seawater. These scenes are not used during HyperSAM training and therefore provide an additional evaluation of transferability to environmental monitoring imagery.

\subsubsection{Baselines}
The comparison includes task-specific methods and recent foundation-model-style methods. For HC, the baselines are SSFTT~\cite{sun2022ssftt}, TGRS-ViT~\cite{zhao2024gscvit}, HyperSIGMA~\cite{wang2024hypersigma}, DOFA~\cite{xiong2024dofa}, HyperFree~\cite{li2025hyperfree}, and SAM3~\cite{carion2025sam3segmentconcepts}. For HAD, we compare with the Reed--Xiaoli detector (RXD)~\cite{reed1990adaptive}, Auto-AD~\cite{mei2022auto}, TDD~\cite{yuan2023onestep}, ADLR~\cite{qu2018adlr}, and the foundation-model baselines. For HCD, we compare with FC-EF/FC-SD~\cite{daudt2018fully}, ML-EDAN~\cite{ml_edan_2022}, SST-Former~\cite{wang2022sstformer}, and the foundation-model baselines. For HTD, the baselines include ACE~\cite{kraut2001adaptive}, the matched filter (MF)~\cite{manolakis2001hyperspectral}, the generalized likelihood ratio test (GLRT)~\cite{kelly1986adaptive}, TSTTD~\cite{jiao2023tsttd}, and the foundation-model baselines.

These baselines are deliberately grouped into classical spectral-statistical detectors, task-specific deep networks, and recent foundation-model-style methods. This grouping helps isolate whether HyperSAM benefits mainly from hyperspectral discrimination, from task-specific optimization, or from the reusable promptable representation learned by the SAM3-based architecture.

 \highlight{The adaptation budget is controlled explicitly. For HC, SSFTT and TGRS-ViT are trained from random initialization using only the one-shot labels. The linear-probing (LP) variants HyperSIGMA-LP and DOFA-LP keep their pretrained backbones frozen and train only a one-shot linear classifier. HyperFree and HyperSAM perform inference without target-scene parameter updates, and the SAM3 baseline uses the same three-channel proxy and prompting protocol as HyperSAM but omits the hyperspectral side encoder and MoE adaptation. All methods use the same seeded support configuration in a given comparison. For HAD, no anomaly labels, target spectra, or manual prompts are provided. For HCD, no changed-pixel labels are used. For HTD, every method receives the same prescribed target spectrum or support prompt. The HyperSIGMA-LP HC values therefore measure a one-shot frozen-backbone linear-probe setting, rather than the task-specific fine-tuning protocol and larger label budgets reported in the original HyperSIGMA study~\cite{wang2024hypersigma}. The comparison should consequently be interpreted as low-label transfer and prompt efficiency, not as a claim about the maximum performance obtainable after full downstream adaptation.}

 \subsubsection{HyperSAM Inference Protocol}
For every test image, the hyperspectral cube is spectrally interpolated to 224 channels and paired with \highlight{the 700/546.1/438.8-nm} three-channel proxy for the frozen RGB branch. The \highlight{original November 2025} \texttt{\highlight{sam3.pt}} \highlight{checkpoint is used for every dataset, without additional natural-image, remote-sensing, or hyperspectral pretraining. The} SAM3 input size is set to $1008\times1008$, and feature level~2 is used for \highlight{mask--feature} matching. Unless otherwise stated, automatic mask generation uses 32 point prompts per side, 64 points per batch, predicted-IoU threshold 0.3, stability threshold 0.4, and box nonmaximum suppression (NMS) threshold 0.7. The HAD experiments use a denser grid with 96 point prompts per side and predicted-IoU threshold 0.4 because anomalies are usually small and can be missed by sparse prompting. Table~\ref{tab:protocol} summarizes the task-specific supervision and decision rules.

\begin{table*}[t]
\centering
\caption{Task-specific inference protocols for HyperSAM. The same model parameters are used for all tasks.}
\label{tab:protocol}
\scriptsize
\renewcommand{\arraystretch}{1.12}
\setlength{\tabcolsep}{2.8pt}
\begin{tabular}{p{0.06\textwidth}|p{0.18\textwidth}|p{0.24\textwidth}|p{0.24\textwidth}|p{0.18\textwidth}}
\hlinew{1pt}
Task & Supervision or prompt & Mask generation & Feature or score construction & Decision rule \\
\hline
HC & One \highlight{randomly sampled} support pixel per class \highlight{from a seeded pool} & 32 points/side; IoU 0.3; stability 0.4; NMS 0.7 & Select the smallest \highlight{covering mask} and average its features\highlight{; use the single-pixel embedding if no mask covers the support} & Assign candidate masks to the nearest prototype by cosine similarity \\
\hline
HAD & No labels, prompts, or target spectra & 96 points/side; IoU 0.4; stability 0.4; NMS 0.7 & Suppress large background masks and retain compact proposals; default area-ratio threshold is 0.0009 & Convert retained mask proposals into an anomaly response map \\
\hline
HCD & No changed-pixel labels & 32 points/side for each date; IoU 0.3; stability 0.4; NMS 0.7 & Average dense features inside masks and compute cosine-distance change scores for both temporal directions & Merge the two directional maps by maximum response and threshold at quantile 0.74 \\
\hline
HTD & One target spectrum or one support prompt & 32 points/side; IoU 0.3; stability 0.4; NMS 0.7 & Locate a target-like support region and use its mean feature as the target prototype & Score candidate masks by cosine similarity to the target prototype and apply light spatial post-processing \\
\hlinew{1pt}
\end{tabular}
\renewcommand{\arraystretch}{1.0}
\end{table*}

 \highlight{For HC, one labeled pixel is drawn uniformly at random from each class without replacement using a seeded generator, and all remaining labeled pixels are reserved for testing. We precompute a pool of 50 one-shot support configurations. The same configuration is used for HyperSAM and every baseline in a given comparison, preventing manual or method-specific support selection. Once the mask-generation and threshold parameters are fixed, HAD and HCD inference are deterministic because they use no target labels. HTD uses the same target spectrum or support prompt for all methods.}

 \begin{table*}[t]
\centering
\caption{Condensed comparison on representative datasets.}
\label{tab:main}
\scriptsize
\setlength{\tabcolsep}{1.75pt}
  \resizebox{\textwidth}{!}{
\begin{tabular}{l|cccccccc|cccccccc}
\hlinew{1pt}
\multicolumn{17}{c}{\textbf{HC} (Indian Pines / PaviaU)} \\
\hline
Metric & SSFTT~\cite{sun2022ssftt} & TGRS-ViT~\cite{zhao2024gscvit} & HyperSIGMA-LP~\cite{wang2024hypersigma} & DOFA-LP~\cite{xiong2024dofa} & HyperFree~\cite{li2025hyperfree} & SpectralEarth~\cite{braham2025spectralearth} & SAM3~\cite{carion2025sam3segmentconcepts} & Ours
& SSFTT~\cite{sun2022ssftt} & TGRS-ViT~\cite{zhao2024gscvit} & HyperSIGMA-LP~\cite{wang2024hypersigma} & DOFA-LP~\cite{xiong2024dofa} & HyperFree~\cite{li2025hyperfree} & SpectralEarth~\cite{braham2025spectralearth} & SAM3~\cite{carion2025sam3segmentconcepts} & Ours \\
\hline
OA (\%)$\uparrow$ & 58.43 & 54.14 & 41.74 & 60.01 & 51.35 & 57.22 & 62.49 & \textbf{69.37}
& 69.34 & 63.99 & 65.27 & 64.22 & 75.23 & 65.48 & 84.61 & \textbf{89.12} \\
AA (\%)$\uparrow$ & 70.07 & 65.28 & 46.99 & 73.64 & 62.07 & 69.90 & 75.80 & \textbf{77.46}
& 71.59 & 75.85 & 72.34 & 63.09 & 62.49 & 56.02 & 85.04 & \textbf{87.38} \\
Kappa (\%)$\uparrow$ & 53.41 & 48.24 & 33.50 & 56.09 & 47.40 & 52.28 & 58.60 & \textbf{65.11}
& 60.65 & 56.01 & 54.67 & 54.36 & 64.53 & 54.57 & 80.45 & \textbf{85.81} \\
\hline
\multicolumn{17}{c}{\textbf{HAD} (Beach-1 / Beach-2)} \\
\hline
Metric & RXD~\cite{reed1990adaptive} & Auto-AD~\cite{mei2022auto} & HyperSIGMA~\cite{wang2024hypersigma} & DOFA~\cite{xiong2024dofa} & TDD~\cite{yuan2023onestep} & HyperFree~\cite{li2025hyperfree} & SpectralEarth~\cite{braham2025spectralearth} & Ours
& RXD~\cite{reed1990adaptive} & Auto-AD~\cite{mei2022auto} & HyperSIGMA~\cite{wang2024hypersigma} & DOFA~\cite{xiong2024dofa} & ADLR~\cite{qu2018adlr} & HyperFree~\cite{li2025hyperfree} & SpectralEarth~\cite{braham2025spectralearth} & Ours \\
\hline
DF$\uparrow$ & 0.9807 & 0.9892 & 0.9948 & 0.9935 & 0.8637 & 0.9255 & 0.9948 & \textbf{0.9988}
& 0.9106 & 0.9460 & 0.8937 & 0.8898 & 0.9081 & 0.9278 & 0.8897 & \textbf{0.9820} \\
ODP$\uparrow$ & 1.2296 & 1.1571 & 1.8943 & 1.8920 & 0.9600 & 1.3510 & \textbf{1.8992} & 1.4977
& 1.0148 & 0.9949 & 1.2677 & 1.1314 & 1.1209 & 1.3556 & 1.1170 & \textbf{1.4641} \\
\hline
\multicolumn{17}{c}{\textbf{HCD} (BayArea/River / Hermiston)} \\
\hline
Metric & FC-EF~\cite{daudt2018fully} & FC-SD~\cite{daudt2018fully} & HyperSIGMA~\cite{wang2024hypersigma} & DOFA~\cite{xiong2024dofa} & ML-EDAN~\cite{ml_edan_2022} & HyperFree~\cite{li2025hyperfree} & SpectralEarth~\cite{braham2025spectralearth} & Ours
& FC-EF~\cite{daudt2018fully} & ML-EDAN~\cite{ml_edan_2022} & HyperSIGMA~\cite{wang2024hypersigma} & DOFA~\cite{xiong2024dofa} & SST-Former~\cite{wang2022sstformer} & HyperFree~\cite{li2025hyperfree} & SpectralEarth~\cite{braham2025spectralearth} & Ours \\
\hline
IoU (\%)$\uparrow$ & 75.78 & \textbf{79.58} & 62.54 & 75.29 & 53.65 & 51.79 & 73.39 & 79.48
& 55.67 & 54.51 & 49.04 & 69.39 & 53.94 & 50.82 & 65.85 & \textbf{70.31} \\
F1 (\%)$\uparrow$ & 86.22 & \textbf{88.63} & 76.95 & 85.90 & 69.83 & 68.24 & 84.65 & 88.57
& 71.52 & 70.55 & 65.81 & 81.93 & 70.08 & 67.39 & 79.41 & \textbf{82.56} \\
\hline
\multicolumn{17}{c}{\textbf{HTD} (Cri / Airport)} \\
\hline
Metric & ACE~\cite{kraut2001adaptive} & MF~\cite{manolakis2001hyperspectral} & HyperSIGMA~\cite{wang2024hypersigma} & DOFA~\cite{xiong2024dofa} & TSTTD~\cite{jiao2023tsttd} & HyperFree~\cite{li2025hyperfree} & SpectralEarth~\cite{braham2025spectralearth} & Ours
& ACE~\cite{kraut2001adaptive} & MF~\cite{manolakis2001hyperspectral} & HyperSIGMA~\cite{wang2024hypersigma} & DOFA~\cite{xiong2024dofa} & GLRT~\cite{kelly1986adaptive} & HyperFree~\cite{li2025hyperfree} & SpectralEarth~\cite{braham2025spectralearth} & Ours \\
\hline
DF$\uparrow$ & 0.9979 & 0.9997 & 0.9978 & 0.9399 & \textbf{0.9999} & 0.9890 & 0.9304 & 0.9975
& 0.6476 & 0.7334 & 0.9070 & 0.7499 & 0.6479 & 0.8876 & 0.8459 & \textbf{0.9149} \\
ODP$\uparrow$ & 1.3043 & 1.4084 & 1.4957 & 1.3798 & \textbf{1.8754} & 1.4828 & 1.2028 & 1.4951
& 0.6720 & 0.7981 & 1.3140 & 0.9998 & 0.6722 & 1.2751 & 1.0460 & \textbf{1.3297} \\
\hlinew{1pt}
\end{tabular}}
 \end{table*}
\subsection{Hyperspectral Classification}

\subsubsection{Experimental Settings}
For HC, HyperSAM is evaluated in a one-shot-per-class setting. On Indian Pines and Pavia University, one labeled support pixel is \highlight{drawn uniformly at random} from each semantic class \highlight{without replacement using a seeded generator, and all} remaining labeled pixels are used for testing. \highlight{A fixed pool of 50 support configurations is precomputed, and the same configuration is used for HyperSAM and all baselines in each comparison.} The model is not fine-tuned on either dataset. HyperSAM first generates masks using the common grid protocol with 32 point prompts per side in Table~\ref{tab:protocol}. For each support pixel, the smallest generated mask covering that pixel is selected to reduce mixed-class context. The class prototype is obtained by averaging the dense spectral-spatial embeddings inside the selected support mask\highlight{. If no generated mask covers a support pixel, its single-pixel embedding is used as a deterministic fallback}. All candidate masks are then assigned to the nearest class prototype by cosine similarity, and the resulting mask labels are converted into a full classification map.

\subsubsection{Results and Analyses}
As shown in Table~\ref{tab:main}, HyperSAM achieves the best OA, AA, and Kappa on both Indian Pines and PaviaU among the compared methods. On Indian Pines, it improves the OA from 62.49\% for SAM3 to 69.37\%, showing that the spectral side branch supplies material information that cannot be recovered from RGB-style priors alone. On PaviaU, HyperSAM reaches 89.12\% OA and 85.81\% Kappa, outperforming HyperFree and SAM3. The comparison with SAM3 is especially informative because both models share the same promptable segmentation prior at inference time. The gain therefore comes from adapting the representation to full-spectrum HSI rather than from changing the prompting protocol. The comparison with HyperFree further indicates that channel-adaptive foundation modeling is not sufficient by itself when the downstream decision relies on object-consistent masks and local spectral separability.

\begin{figure*}[t]
\centering
\includegraphics[width=0.98\textwidth]{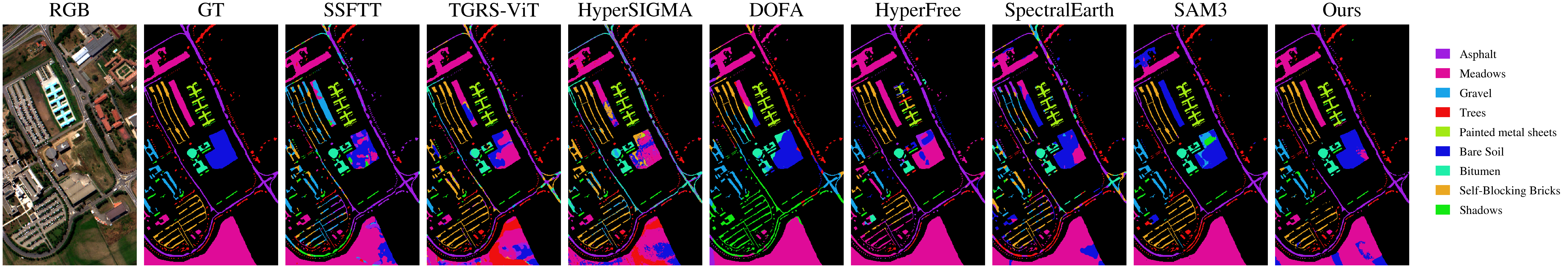}
\caption{Qualitative comparison of hyperspectral classification results. Compared to the baselines, HyperSAM produces more spatially coherent classification maps with sharper land-cover boundaries and significantly less salt-and-pepper noise, demonstrating the benefit of region-level promptable priors.}
\label{fig:hc_vis}
\end{figure*}

The one-shot protocol also reveals why the smallest-mask support rule is useful. A single labeled pixel may lie inside a mixed agricultural parcel or near an urban boundary. Selecting the smallest generated mask that covers the support point reduces the amount of irrelevant context used to form the class prototype, which directly matches the data-quality argument in Section~\ref{sec:data}: object-centric masks are valuable because they turn sparse labels into region-level but still boundary-aware supervision. We further provide a qualitative comparison in Fig.~\ref{fig:hc_vis}. Consistent with the quantitative metrics, traditional and pixel-level foundation-model baselines often suffer from salt-and-pepper noise or fragmented predictions within homogeneous regions. In contrast, HyperSAM produces highly coherent classification maps with sharp and accurate land-cover boundaries. This visual improvement highlights the complementary nature of our design: the promptable object priors provide coherent spatial support that suppresses local noise, while the full-spectrum features ensure accurate class discrimination in spectrally subtle areas.

\subsection{Hyperspectral Anomaly Detection}

\subsubsection{Experimental Settings}
HAD is evaluated on Beach-1 and Beach-2 from the Airport-Beach-Urban benchmark~\cite{tu2019sdbp}. No target spectrum, support prompt, or anomaly label is used during inference. HyperSAM generates dense mask proposals using 96 point prompts per side, 64 points per batch, predicted-IoU threshold 0.4, stability threshold 0.4, and box NMS threshold 0.7. Because the anomalies in these scenes are spatially compact, large masks are treated as background structures and suppressed. Candidate masks with relative area below the default threshold 0.0009 are retained as anomaly proposals, and their mask-level responses are projected back to the pixel grid to produce the final anomaly map.

\subsubsection{Results and Analyses}
Table~\ref{tab:main} shows that HyperSAM obtains the best DF on both Beach-1 and Beach-2 and the best ODP on Beach-2. On Beach-1, HyperSIGMA and DOFA obtain higher ODP because their score distributions are more aggressive around the target, but HyperSAM still gives the highest DF and cleaner object-level localization. This distinction is important: a high DF indicates that the compact abnormal object is captured by the promptable mask proposals, whereas ODP is also affected by score calibration and background ranking. HyperSAM therefore does not simply amplify all high-contrast pixels. It first searches for spatially coherent small regions and then converts them into an anomaly map.

\begin{figure*}[t]
\centering
\includegraphics[width=0.98\textwidth]{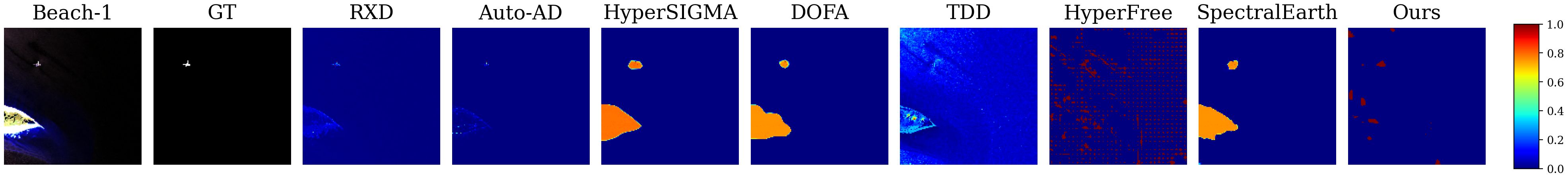}
\caption{Qualitative comparison on the Beach-1 anomaly detection scene from the Airport-Beach-Urban benchmark~\cite{tu2019sdbp}. HyperSAM localizes the target with compact mask-level responses while suppressing most background structures.}
\label{fig:had_beach1}
\end{figure*}

\begin{figure*}[t]
\centering
\includegraphics[width=0.98\textwidth]{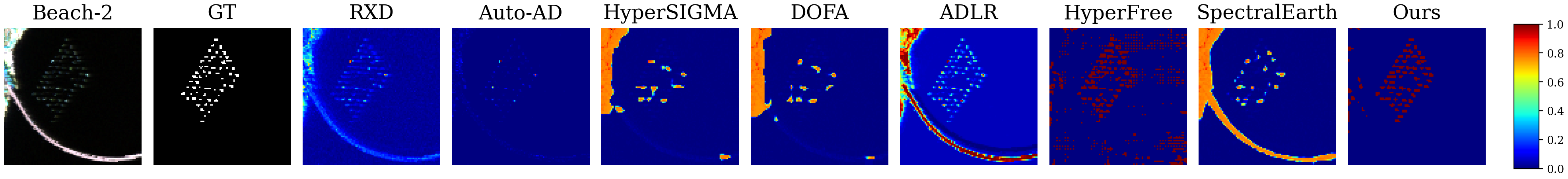}
\caption{Qualitative comparison on the Beach-2 anomaly detection scene from the Airport-Beach-Urban benchmark~\cite{tu2019sdbp}. HyperSAM reduces false alarms from shoreline and water-background regions and highlights coherent anomalous targets.}
\label{fig:had_beach2}
\end{figure*}

The qualitative comparisons in Fig.~\ref{fig:had_beach1} and Fig.~\ref{fig:had_beach2} show the same trend: classical and task-specific detectors often activate ship edges, water boundaries, or background structures, whereas HyperSAM produces compact responses around anomalous objects with fewer scattered false alarms. This behavior supports the motivation for constructing high-resolution pseudo-mask supervision. Even though HAD is unsupervised at test time, the learned ability to propose reliable small regions helps transform pixel-level spectral abnormality into object-level evidence, which is difficult to obtain from reconstruction-only foundation features.

\subsection{Hyperspectral Change Detection}

\subsubsection{Experimental Settings}
HCD is evaluated on BayArea/River and Hermiston following common hyperspectral change-detection protocols~\cite{ml_edan_2022,wang2022sstformer}. HyperSAM is used in a zero-shot setting: no changed or unchanged pixels from the target pair are used to train a classifier or tune the model. For each bi-temporal pair, masks are generated independently on the two dates using 32 point prompts per side, predicted-IoU threshold 0.3, stability threshold 0.4, and box NMS threshold 0.7. For a mask generated at one date, HyperSAM averages the dense embeddings inside the same spatial region at both dates and computes a cosine-distance change score. This produces two directional change maps, one from each temporal mask set~\cite{wei2026beyond}. The final score map is obtained by taking the maximum response of the two directions, and the binary change map is produced using the fixed quantile threshold 0.74.

\subsubsection{Results and Analyses}
On BayArea/River, HyperSAM is close to the task-specific FC-SD detector~\cite{daudt2018fully} and outperforms the other foundation-model baselines. On Hermiston, HyperSAM achieves the best IoU and F1 among all compared methods, improving IoU from 50.82\% for HyperFree~\cite{li2025hyperfree} to 70.31\%. The improvement is consistent with the prior-utilization bottleneck discussed in the introduction: change detection requires not only spectral sensitivity, but also a stable spatial unit over which spectral differences can be compared. Pixel-level distances can be unstable under slight misregistration, seasonal variation, or local illumination differences, while the mask-level comparison averages features inside coherent regions and therefore reduces local noise.

\begin{figure*}[t]
\centering
\includegraphics[width=0.98\textwidth]{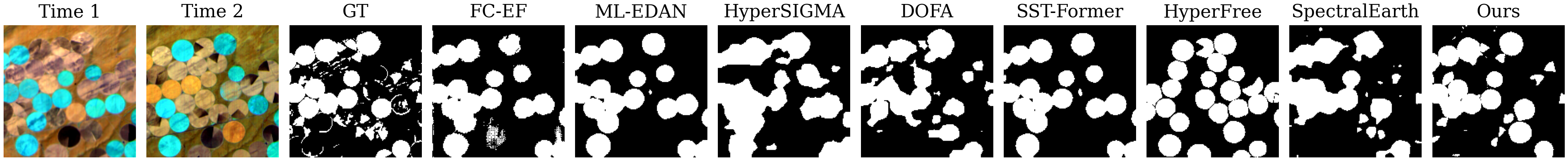}
\caption{Qualitative comparison on the Hermiston change detection scene~\cite{wang2022sstformer}. HyperSAM produces more complete changed regions and fewer fragmented artifacts than the compared foundation-model baselines.}
\label{fig:hcd_hermiston}
\end{figure*}

Fig.~\ref{fig:hcd_hermiston} further shows that HyperSAM better preserves the shapes of changed regions than HyperFree and avoids the heavy fragmentation observed in some pixel-level or weakly adapted baselines. The two-directional mask comparison also matters. Masks generated from the first date may better describe pre-change objects, while masks generated from the second date may better describe newly emerged or disappeared structures. Taking the maximum of both directional responses provides a simple way to reuse the same promptable interface for temporal reasoning without introducing a separate change-detection head.

\subsection{Hyperspectral Target Detection}

\subsubsection{Experimental Settings}
HTD is evaluated on the Nuance Cri scene~\cite{zhang2017ajsrm} and the Airport scene from the Airport-Beach-Urban benchmark~\cite{tu2019sdbp}. Each test scene provides one target spectrum or one target support prompt. When a target spectrum is available, the spectrum is first matched with the normalized hyperspectral cube by cosine similarity, and the highest-response location is used as a target support point. HyperSAM then generates candidate masks using 32 point prompts per side, predicted-IoU threshold 0.3, stability threshold 0.4, and box NMS threshold 0.7. The generated mask containing the support point is used to compute a region-level target prototype from dense embeddings. All candidate masks are scored by cosine similarity to this target prototype\highlight{. Fixed cosine thresholds of 0.73 for Cri and 0.61 for Airport are then applied, followed by spatial dilation with a $3\times3$ kernel to produce} the final detection \highlight{maps}.

\subsubsection{Results and Analyses}
On Cri, specialized spectral detectors are extremely strong, and TSTTD~\cite{jiao2023tsttd} achieves the best DF and ODP. HyperSAM remains competitive, with DF close to classical ACE~\cite{kraut2001adaptive} and MF~\cite{manolakis2001hyperspectral}. This is a useful negative case for a unified model: when the target is almost perfectly defined by a distinctive spectrum and little spatial ambiguity exists, a carefully designed spectral detector can still be preferable. HyperSAM should therefore be interpreted as a transferable promptable model rather than a replacement for every specialized detector in every sensing condition.

\begin{figure*}[t]
\centering
\includegraphics[width=0.98\textwidth]{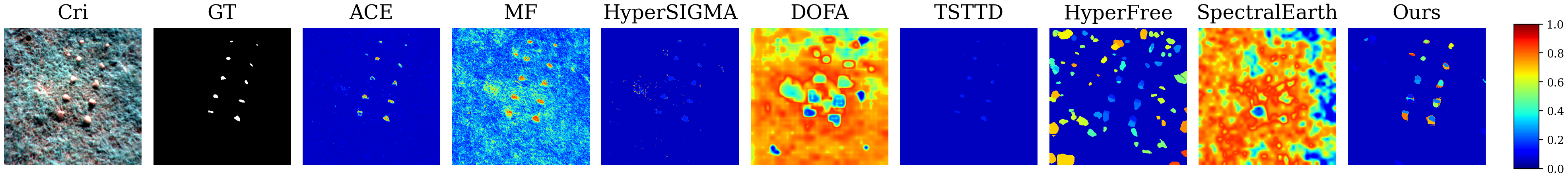}
\caption{Qualitative comparison on the Nuance Cri target detection scene~\cite{zhang2017ajsrm}. HyperSAM remains competitive with strong spectral detectors while producing region-consistent target responses.}
\label{fig:htd_cri}
\end{figure*}

\begin{figure*}[t]
\centering
\includegraphics[width=0.98\textwidth]{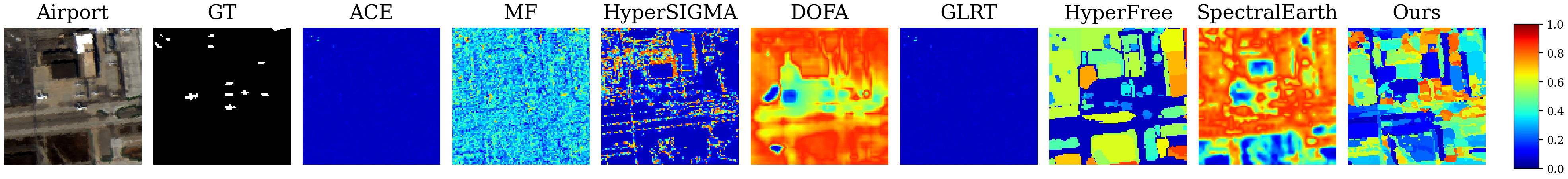}
\caption{Qualitative comparison on the Airport target detection scene from the Airport-Beach-Urban benchmark~\cite{tu2019sdbp}. HyperSAM benefits from object-level mask features and achieves stronger clutter suppression than several spectral or foundation-model baselines. \highlight{The false-color composite uses the closest available bands to 700, 546.1, and 438.8~nm as red, green, and blue, respectively.}}
\label{fig:htd_airport}
\end{figure*}

On Airport, where spatial context and object-level grouping are more important, HyperSAM achieves the best DF and ODP, outperforming HyperFree~\cite{li2025hyperfree}, HyperSIGMA~\cite{wang2024hypersigma}, and DOFA~\cite{xiong2024dofa}. The visual results in Fig.~\ref{fig:htd_cri} and Fig.~\ref{fig:htd_airport} clarify this behavior.  In the compact Cri scene, purely spectral matching is already highly effective, while HyperSAM still produces target-focused responses with limited background activation. In the more structured Airport scene, region-level mask features help suppress clutter and aggregate target evidence over object-like areas. This task-dependent behavior strengthens the cross-task message: the benefit of promptable spectral-spatial modeling is largest when spectral evidence and object grouping are complementary.

\subsection{Ablation Studies and Further Analysis}

\subsubsection{Data Construction\highlight{, Architecture Components,} and MoE Refinement}
Table~\ref{tab:ablation} \highlight{explicitly separates the training corpus, frozen RGB prior, hyperspectral side encoder, MoE refinement, and confidence-aware weighting}. The HyperFree-data setting follows the larger but less object-centric Hyper-Seg corpus~\cite{li2025hyperfree}, whereas \highlight{the synthetic-HSI settings use the physically constrained SpaceNet-derived corpus. The comparison shows} that sample count alone is not sufficient for promptable hyperspectral learning\highlight{: spectral} plausibility, spatial sharpness, semantic diversity, and pseudo-mask reliability are \highlight{especially} important because mask supervision \highlight{depends on} object boundaries and region consistency. \highlight{Raw multispectral input} remains competitive on PaviaU HC, \highlight{indicating} that directly observed \highlight{bands can capture} scene-specific land-cover cues\highlight{, whereas the synthesized full spectrum is more useful when material} shape and target/background separability are \highlight{central.}

\highlight{The removal experiments further show that the two branches are complementary. Removing the frozen RGB branch weakens the inherited objectness, boundary, and prompt-response priors, while removing the hyperspectral side encoder eliminates material-sensitive} full-spectrum \highlight{corrections. The MoE path improves scale-adaptive mask refinement, and confidence-aware weighting increases robustness to structured pseudo-label noise. Thus, these components address distinct parts of the adaptation problem rather than duplicating the same function.}

\begin{table*}[t]
\centering
\caption{\highlight{Component-wise ablation of HyperSAM. Components are grouped into training data, network structure, and confidence-aware training.}}
\label{tab:ablation}
\scriptsize
\setlength{\tabcolsep}{2.0pt}
\renewcommand{\arraystretch}{1.08}
\resizebox{\textwidth}{!}{%
\begin{tabular}{l|ccc|ccc|c|cccc}
\hlinew{1pt}
\multirow{3}{*}{Configuration}
& \multicolumn{7}{c|}{Component}
& \multicolumn{4}{c}{Downstream performance} \\
\cline{2-12}
& \multicolumn{3}{c|}{Training data}
& \multicolumn{3}{c|}{Network structure}
& \multicolumn{1}{c|}{Training loss}
& \multirow{2}{*}{PaviaU OA$\uparrow$}
& \multirow{2}{*}{Beach-2 ODP$\uparrow$}
& \multirow{2}{*}{Hermiston IoU$\uparrow$}
& \multirow{2}{*}{Airport ODP$\uparrow$} \\
\cline{2-8}
& HyperFree & Raw-MS & Syn-HSI & RGB & HSI & MoE & Conf. & & & & \\
\hline
HyperFree training data
& \checkmark & -- & -- & \checkmark & \checkmark & \checkmark & \checkmark
& 79.33 & 1.3903 & 68.40 & 1.2500 \\
Raw multispectral input
& -- & \checkmark & -- & \checkmark & \checkmark & \checkmark & \checkmark
& 85.84 & 1.3576 & 66.73 & 1.3088 \\
Synthetic hyperspectral input
& -- & -- & \checkmark & \checkmark & \checkmark & -- & \checkmark
& 84.92 & 1.4452 & 66.61 & 1.3544 \\
Without frozen RGB branch
& -- & -- & \checkmark & -- & \checkmark & \checkmark & \checkmark
& 84.89 & 1.3126 & 66.45 & 1.3208 \\
Without hyperspectral side encoder
& -- & -- & \checkmark & \checkmark & -- & \checkmark & \checkmark
& 84.61 & 1.2948 & 64.96 & 1.3073 \\
Synthetic HSI + MoE
& -- & -- & \checkmark & \checkmark & \checkmark & \checkmark & --
& 87.48 & 1.4524 & 68.96 & \textbf{1.3587} \\
Full HyperSAM
& -- & -- & \checkmark & \checkmark & \checkmark & \checkmark & \checkmark
& \textbf{89.12} & \textbf{1.4641} & \textbf{70.31} & 1.3297 \\
\hlinew{1pt}
\end{tabular}}
\renewcommand{\arraystretch}{1.0}
\end{table*}

\highlight{Table~\ref{tab:initialization_ablation} isolates the prior-preserving initialization while holding the data, frozen RGB branch, side encoder, MoE path, optimization, and evaluation protocol fixed. Randomly initialized injection modules can perturb the pretrained RGB features at the beginning of training. Zero initialization instead begins with zero residual injection, so the network initially reproduces the frozen SAM3 representation and introduces hyperspectral corrections gradually. The consistent improvement supports this stable adaptation strategy.}

\begin{table}[t]
\centering
 \caption{\highlight{Ablation of the feature-injection initialization strategy. All other settings are unchanged.}}
\label{tab:initialization_ablation}
\scriptsize
\setlength{\tabcolsep}{3.0pt}
\renewcommand{\arraystretch}{1.04}
\resizebox{0.48\textwidth}{!}{%
\begin{tabular}{l|cccc}
\hlinew{1pt}
Initialization & PaviaU OA$\uparrow$ & Beach-2 ODP$\uparrow$ & Hermiston IoU$\uparrow$ & Airport ODP$\uparrow$ \\
\hline
Random & 83.59 & 1.3967 & 68.43 & 1.3126 \\
Zero & \textbf{89.12} & \textbf{1.4641} & \textbf{70.31} & \textbf{1.3297} \\
\hlinew{1pt}
\end{tabular}}
\renewcommand{\arraystretch}{1.0}
\end{table}

\highlight{We also conduct a controlled leave-one-term-out ablation of the reconstruction objective in Eq.~\eqref{eq:recon_loss}. Every variant uses the same paired spectral library, SpaceNet patches, initialization, optimizer, learning-rate schedule, and number of iterations. Only the indicated term is removed. The same HyperSAM configuration is then trained on each resulting corpus and evaluated with unchanged downstream protocols. As shown in Table~\ref{tab:generator_loss_ablation}, the complete objective achieves the best overall transfer. Removing $L_1$ weakens radiometric fidelity, removing $L_{\cos}$ reduces spectral-shape consistency, removing $L_{\mathrm{proj}}$ weakens agreement with the observed WorldView-3 signal, removing $L_{\mathrm{sp}}$ produces less parsimonious material mixtures, and removing $L_{\mathrm{tv}}$ reduces spatial coherence in abundance and illumination maps. The results indicate complementary rather than redundant constraints.}

\begin{table}[t]
\centering
 \caption{ \highlight{Leave-one-term-out ablation of the hyperspectral data-generation objective. All variants use the same training data}, \highlight{initialization}, and \highlight{optimization schedule}.}
  \label{tab:generator_loss_ablation}
 \scriptsize
  \setlength{\tabcolsep}{4.0pt}
\renewcommand{\arraystretch}{1.08}
\resizebox{\linewidth}{!}{%
\begin{tabular}{l|cccc}
\hlinew{1pt}
Configuration & PaviaU OA$\uparrow$ & Beach-2 ODP$\uparrow$ & Hermiston IoU$\uparrow$ & Airport ODP$\uparrow$ \\
\hline
w/o $L_1$ reconstruction & 81.64 & 1.3846 & 63.88 & 1.2715 \\
w/o cosine consistency $L_{\cos}$ & 80.26 & 1.2681 & 61.15 & 1.2194 \\
w/o projection consistency $L_{\mathrm{proj}}$ & 87.91 & 1.4218 & 68.74 & 1.3312 \\
w/o abundance sparsity $L_{\mathrm{1/2}}$ & 83.52 & 1.3467 & 65.21 & 1.2916 \\
w/o total variation $L_{\mathrm{tv}}$ & 88.73 & 1.4479 & 69.42 & \textbf{1.3421} \\
Full generator objective & \textbf{89.12} & \textbf{1.4641} & \textbf{70.31} & 1.3297 \\
\hlinew{1pt}
\end{tabular}}
 \renewcommand{\arraystretch}{1.0}
  \end{table}

\begin{figure}[t]
\centering
\includegraphics[width=\linewidth]{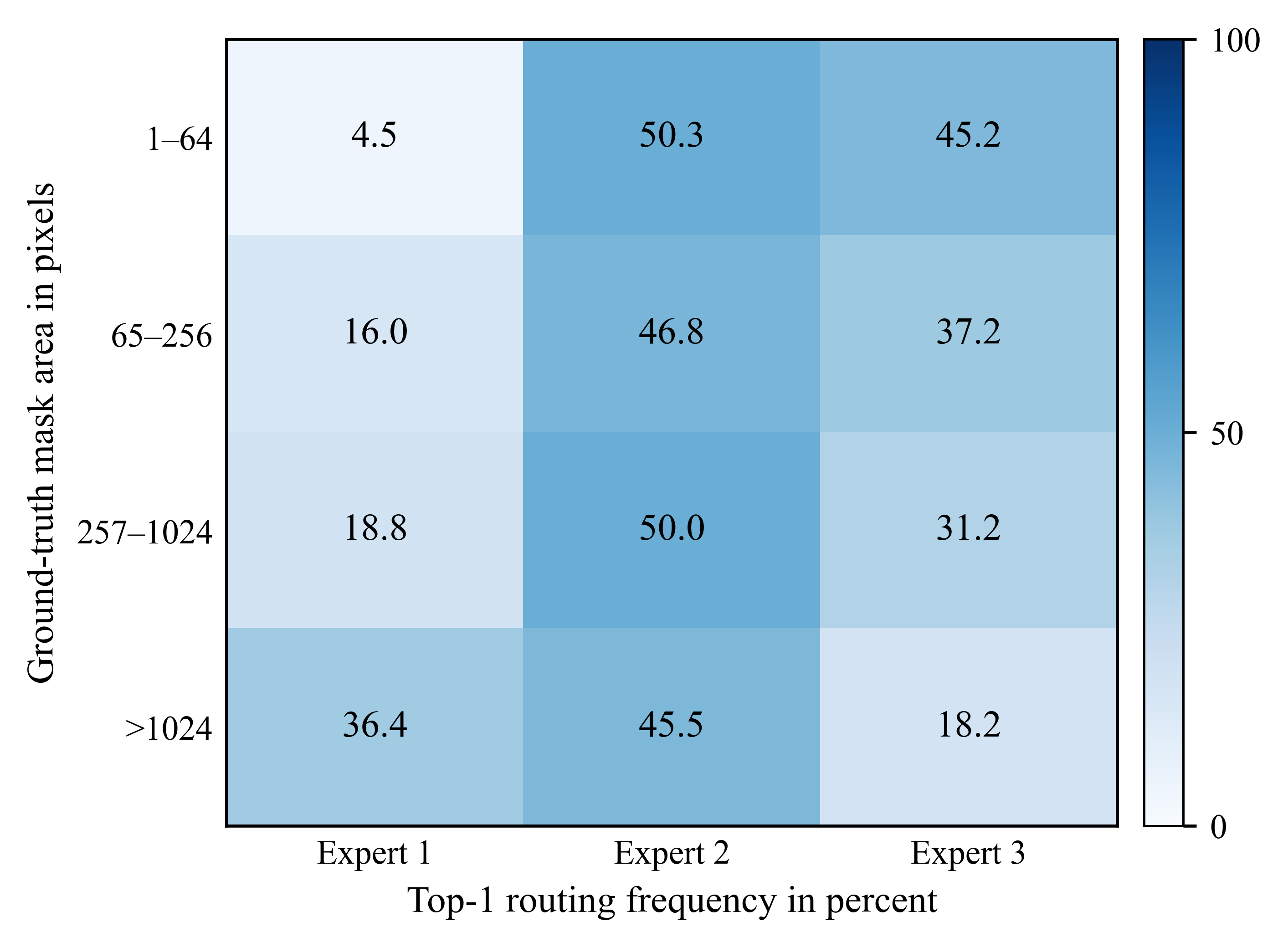}
\caption{\highlight{Top-1 routing frequency of the three MoE mask-refinement experts across mask-size ranges, showing scale-dependent specialization without expert collapse.}}
\label{fig:moe_routing}
\end{figure}

\highlight{The MoE routing statistics in Fig.~\ref{fig:moe_routing} provide a direct interpretation of the refiner. Expert~3 is selected for 45.2\% of masks containing 1--64 pixels but only 18.2\% of masks larger than 1024 pixels, whereas Expert~1 increases from 4.5\% for the smallest masks to 36.4\% for the largest. Expert~2 remains active across all ranges, with routing frequencies from 45.5\% to 50.3\%. This soft division of labor indicates that Expert~3 specializes more strongly in small-object refinement, Expert~1 becomes increasingly important for large regions, and Expert~2 captures scale-shared patterns without expert collapse.}

\subsubsection{Confidence-Aware Noisy-Label Weighting}
\begin{figure}[t]
\centering
\includegraphics[width=1\linewidth]{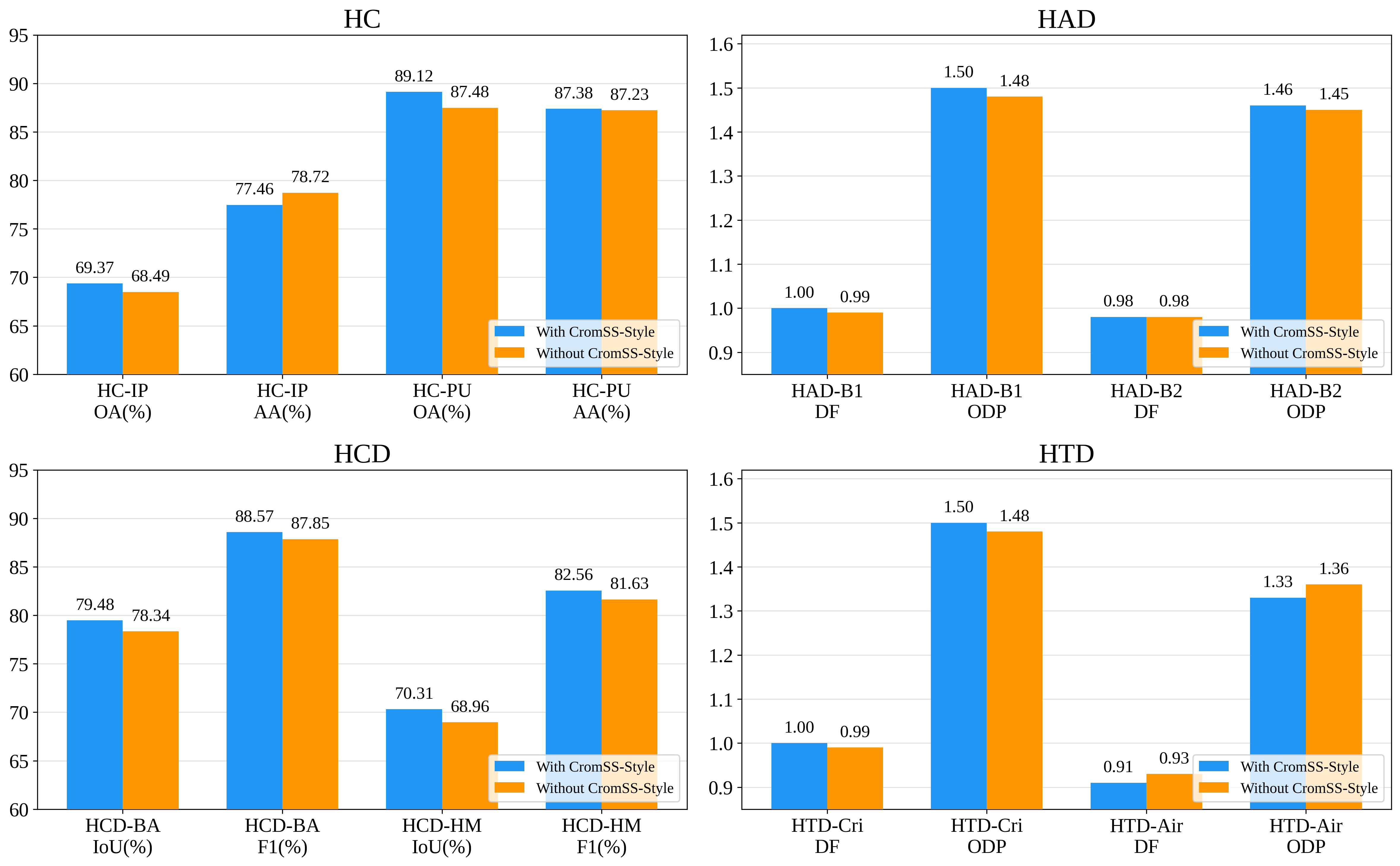}
\caption{Ablation of CromSS-style confidence-aware robust learning~\cite{liu2025cromss}. The comparison reports representative HC, HAD, HCD, and HTD metrics with and without noisy-label weighting and cross-view consistency.}
\label{fig:ablation_cromss}
\end{figure}

Fig.~\ref{fig:ablation_cromss} compares HyperSAM with and without the CromSS-style confidence-aware robust learning~\cite{liu2025cromss}. The robust version improves the main HC result on PaviaU, the HCD results on both BayArea/River and Hermiston, and the HAD ODP scores. This confirms that pseudo-mask supervision benefits from selecting reliable regions and softly down-weighting uncertain boundaries, holes, and ambiguous background leakage. The result also clarifies a common issue in hyperspectral foundation-model training: pseudo-masks are not uniformly wrong or uniformly correct. Interior regions of roads, buildings, water, or vegetation often provide useful supervision, whereas thin boundaries, shadows, mixed pixels, and small objects introduce structured label noise. Treating these two cases differently is more appropriate than discarding pseudo-masks altogether.

The HTD results are more mixed, especially on Airport, which indicates that target detection is sensitive to prototype selection, score calibration, and thresholding. Thus, confidence-aware weighting mainly improves the reliability of mask representation learning rather than replacing task-specific detection calibration. To illustrate this mechanism, Fig.~\ref{fig:weights_vis} visualizes the learned confidence weight map ($W$). As shown in the figure, the network successfully assigns higher weights to structurally clear and reliable semantic regions. In contrast, ambiguous object boundaries and complex background fragments receive significantly lower weights. This qualitative observation directly validates our robust learning design: it prevents the model from blindly fitting noisy pseudo-labels at object borders while preserving the promptable foundation-model priors inherited from the frozen RGB branch.

 \highlight{Table~\ref{tab:training_loss_ablation} further isolates the optimization components while keeping the model, training corpus, initialization, schedule, and downstream protocols fixed. BCE and Dice provide complementary pixelwise and region-overlap supervision. Confidence weighting suppresses unreliable pseudo-label pixels. $\mathcal{L}_{\mathrm{cons}}$ improves invariance across spectral windows, and $\mathcal{L}_{\mathrm{bal}}$ prevents expert collapse. The complete objective gives the best overall HC, HAD, and HCD performance, while the Airport HTD result again shows that specialized threshold calibration can trade off against representation quality.}

\begin{table}[t]
\centering
\caption{\highlight{Ablation of the HyperSAM training-loss components. All variants use the same architecture, corpus, initialization, optimization schedule, and downstream protocol.}}
\label{tab:training_loss_ablation}
\scriptsize
\setlength{\tabcolsep}{4.2pt}
\renewcommand{\arraystretch}{1.08}
\resizebox{\linewidth}{!}{%
\begin{tabular}{l|cccc}
\hlinew{1pt}
Configuration & PaviaU OA$\uparrow$ & Beach-2 ODP$\uparrow$ & Hermiston IoU$\uparrow$ & Airport ODP$\uparrow$ \\
\hline
w/o Dice & 85.06 & 1.4159 & 67.16 & 1.3226 \\
w/o BCE & 86.24 & 1.4273 & 67.81 & 1.3396 \\
w/o confidence weighting & 87.48 & 1.4524 & 68.96 & \textbf{1.3587} \\
w/o $\mathcal{L}_{\mathrm{bal}}$ & 87.17 & 1.4592 & 68.27 & 1.3472 \\
Full training objective & \textbf{89.12} & \textbf{1.4641} & \textbf{70.31} & 1.3297 \\
\hlinew{1pt}
\end{tabular}}
\renewcommand{\arraystretch}{1.0}
\end{table}

\subsubsection{Cross-Task Summary}
The experiments show that HyperSAM is strongest when object-level grouping and spectral-spatial features are both useful. HC and HCD benefit from mask-level features because they require coherent region interpretation rather than isolated pixel decisions. HAD benefits from dense promptable proposals because small coherent anomalies can be separated from large background regions. HTD is more scene-dependent: classical spectral detectors such as ACE, MF, GLRT, and TSTTD~\cite{kraut2001adaptive,manolakis2001hyperspectral,kelly1986adaptive,jiao2023tsttd} can dominate when the target is defined almost perfectly by its spectrum, whereas HyperSAM is advantageous when spatial context and object grouping help suppress clutter.

The \highlight{expanded ablations} provide the same message from another angle. \highlight{Improvements in} HC, HAD, and HCD \highlight{generally track the quality of abundance-transfer data construction, prior-preserving spectral adaptation, MoE mask refinement,} and \highlight{confidence-aware learning, whereas HTD is more sensitive to the target prototype, score calibration, and thresholding. The} most transferable parts of HyperSAM are \highlight{therefore the spectral-mask corpus, frozen-prior} adaptation, and \highlight{robust mask representation, while specialized detection can still benefit from} task-aware calibration\highlight{. Within the evaluated one-shot, zero-shot, and small-scene protocols, these results support} HyperSAM as a unified promptable spectral-spatial model rather than a \highlight{collection} of task-specific networks\highlight{.}

\highlight{The current study also has important limitations. HyperSAM is trained exclusively on synthesized hyperspectral data, and measured endmembers plus multispectral reconstruction constraints cannot reproduce the full variability of real acquisitions. Differences in sensor response, atmosphere, spatial resolution, geography, season, illumination, and land-cover distribution introduce a synthetic-to-real domain gap. SAM3-derived pseudo-masks can disagree with hyperspectral material boundaries, particularly around shadows, thin structures, mixed pixels, and spectrally distinct but visually similar surfaces. The method also depends on the availability and stability of the specific original SAM3 checkpoint used here. Finally, the present benchmarks emphasize frozen-checkpoint, one-shot}, \highlight{zero-shot, or small-scene inference. They do not establish superiority under large-patch training or full downstream fine-tuning, and specialized spectral detectors can remain preferable when a target is almost completely defined by its spectrum. Future work should combine diverse real annotated HSI with the synthetic corpus, broaden sensor and material coverage, and evaluate large-patch} and \highlight{full-fine-tuning regimes}.

\subsection{Real-World Applicability}

  \subsubsection{Experimental Settings}
  We further evaluate HyperSAM on the HOSD Gulf of Mexico GM13/GM17 scenes~\cite{duan2023hosd}. The experiment is formulated as binary classification between oil spill and clean seawater. Oil-spill mapping is included because it stresses the necessity of hyperspectral sensing in a practical environmental-monitoring scenario: SAR and RGB observations can capture slick-like structures, but look-alike sea surfaces, illumination changes, and wave-related dark regions can make purely spatial or intensity-based cues ambiguous~\cite{brekke2005oilspill}. Dense hyperspectral measurements provide additional material-level evidence for separating oil-covered water from clean water and non-oil sea surfaces~\cite{duan2023hosd}. Each original 224-band AVIRIS cube is directly processed by the common HyperSAM spectral interface without task-specific fine-tuning. We randomly sample 50 labeled pixels from each class to construct the corresponding spectral-spatial prototypes, while all remaining pixels are used for evaluation. The trained checkpoint and automatic mask generation parameters remain unchanged.

  \subsubsection{Results and Analyses}
HyperSAM achieves an OA of 91.14\% and an AA of 90.83\% on HOSD-GM13, and an OA of 97.39\% and an AA of 97.65\% on HOSD-GM17. These results indicate that HyperSAM captures useful representations for both oil-covered and clean-water regions despite the substantial difference between these marine scenes and the urban synthetic imagery used during training. The experiment therefore provides a complementary form of validation: the model is not only competitive on standard HC/HAD/HCD/HTD benchmarks, but can also transfer its promptable spectral-spatial representation to an application where material discrimination is essential.

\begin{figure}[t]
\centering
\includegraphics[width=\linewidth]{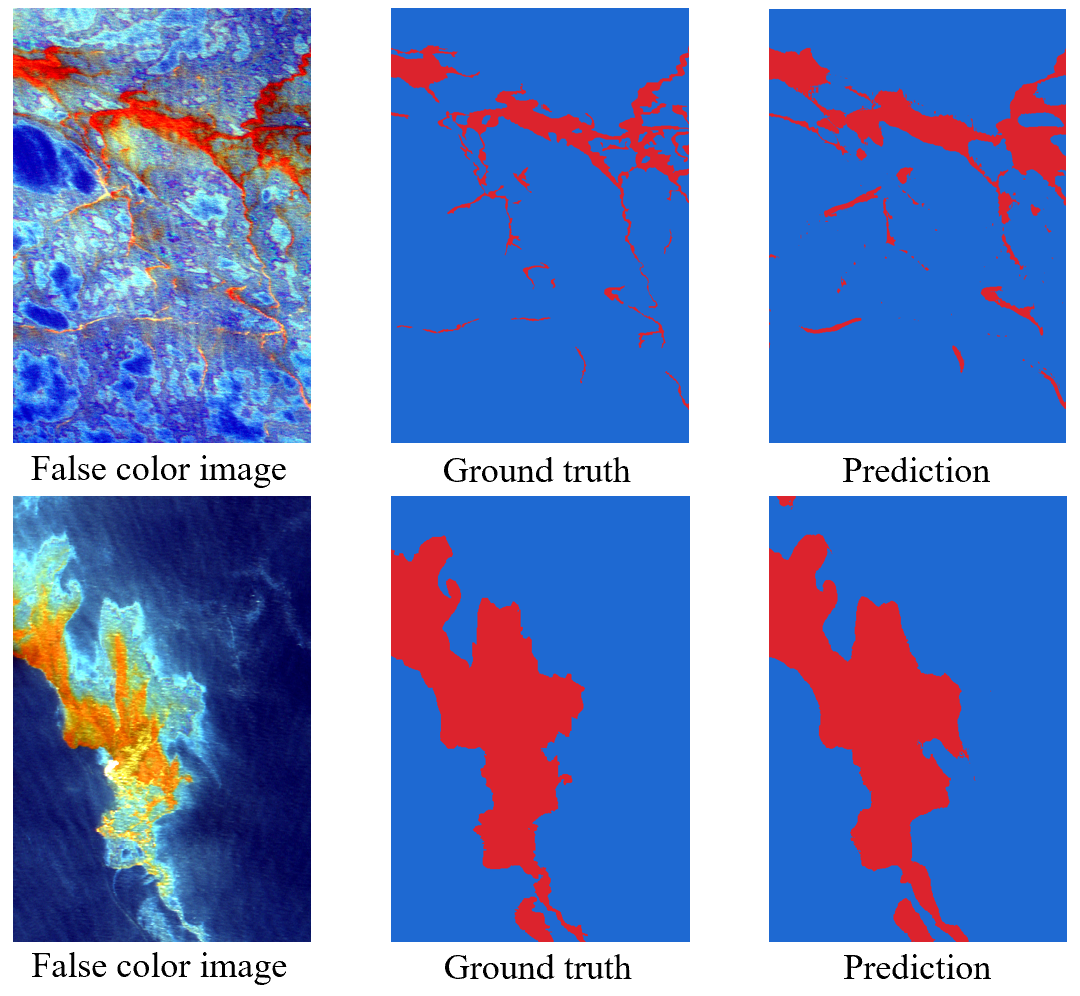}
\caption{Oil-spill mapping results on the HOSD Gulf of Mexico GM13  and GM17 \highlight{scenes}. From left to right, the figure shows the input \highlight{false-color composite}, ground truth, and HyperSAM prediction for each scene. \highlight{The composites use the closest available bands to 700, 546.1, and 438.8~nm as red, green, and blue, respectively.} Red and blue \highlight{in the label maps} denote oil spill and background (clean seawater or non-oil sea surface), respectively.}
\label{fig:oilspill}
\end{figure}

As shown in Fig.~\ref{fig:oilspill}, the predictions preserve large-scale oil-spill structures, although errors remain around fragmented slick boundaries and spectrally ambiguous transition regions. These failure cases are meaningful because they expose the remaining gap between synthetic urban training data and complex marine scenes. They also suggest a practical extension of the proposed data pipeline: if additional water, coastline, and oil-like spectral endmembers are incorporated, the same abundance-transfer strategy could synthesize more application-specific spectral-mask pairs for environmental monitoring without changing the HyperSAM architecture.

\subsection{Computational Cost and Carbon Footprint}
\label{sec:carbon}

In addition to predictive performance, we evaluate the operational cost of
deploying different hyperspectral classification pipelines under the same
one-shot Indian Pines protocol. Reporting energy and carbon cost alongside
accuracy has become increasingly important for large neural models and
foundation-model research~\cite{strubell2019energy,anthony2020carbontracker,patterson2021carbon}. The benchmark is designed to measure the
target-scene training, adaptation, and inference cost, rather than the much
larger and method-dependent cost of upstream foundation-model pretraining.
Each method is executed on a single NVIDIA RTX 4090 GPU three
times from a cold start. For each repetition, one support configuration is
selected from a fixed pool of 50 precomputed one-shot support configurations,
so that the reported accuracy and cost statistics are tied to the same
executed runs. For HyperSAM and HyperFree, which do not update parameters on
the target scene, the measured command includes checkpoint loading, automatic
mask generation, dense feature extraction, support-prototype construction,
and full-scene classification. For SSFTT and TGRS-ViT, the measured command
includes one-shot supervised training from random initialization followed by
full-scene inference. For HyperSIGMA-LP and DOFA-LP, it includes frozen
backbone feature extraction, one-shot linear-probe fitting, and full-scene
inference.

\begin{figure}[t]
    \centering
    \includegraphics[width=\linewidth]{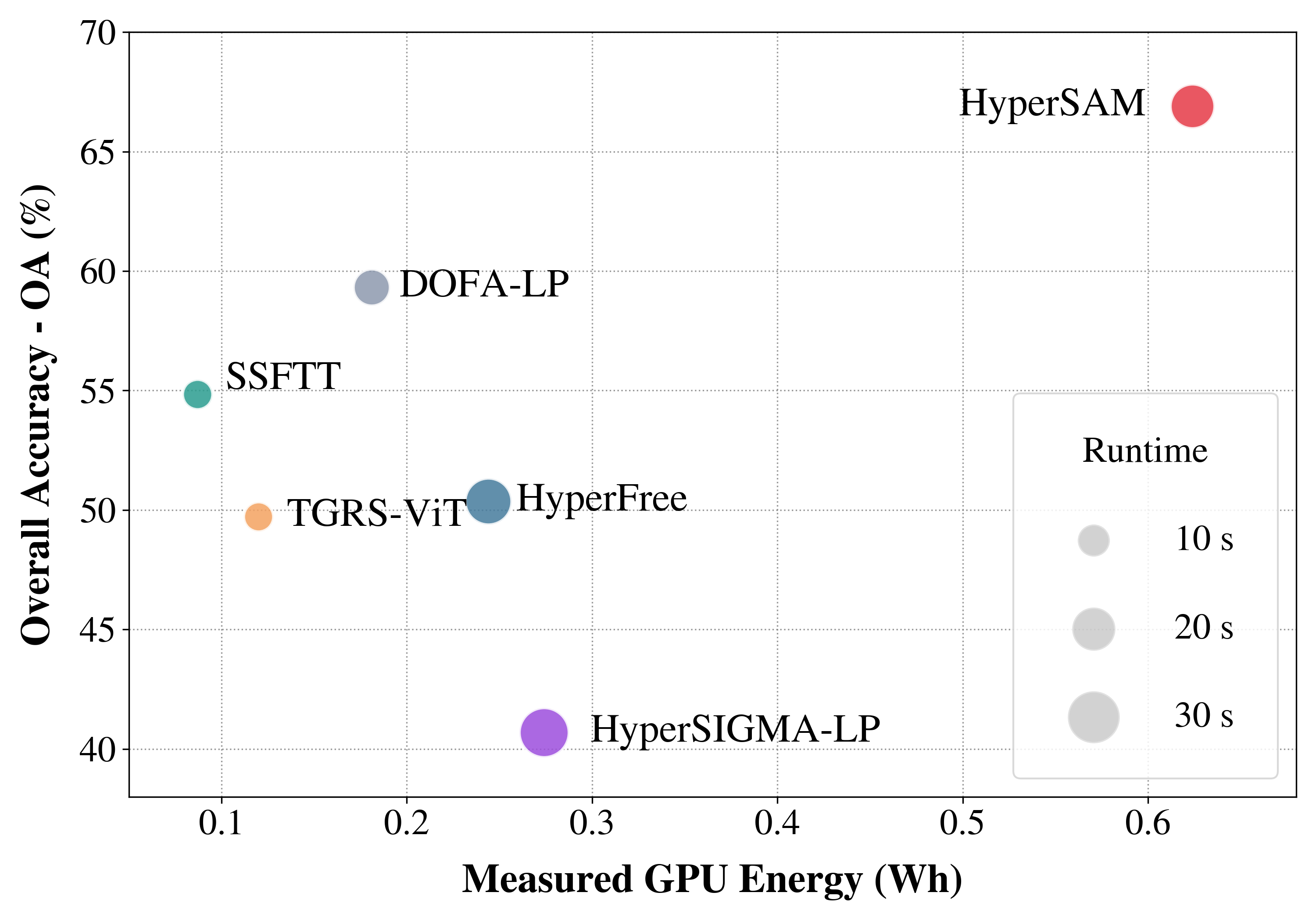}
    \caption{Accuracy--energy trade-off on Indian Pines. The horizontal axis reports measured GPU board energy for downstream
    training, adaptation, and/or inference. }
    \label{fig:carbon_efficiency}
\end{figure}

GPU board power is monitored at a sampling interval
of 0.5~s. Each run is accepted only when no pre-existing GPU compute process
is detected, in order to avoid contamination from other workloads. The
sampled board-power sequence is integrated by the trapezoidal rule to obtain
the measured GPU energy. A short idle-power sample is also recorded by the
measurement script for diagnostic purposes and for internally saved
incremental-energy estimates. However, the values reported in
Table~\ref{tab:carbon} and Fig.~\ref{fig:carbon_efficiency} use the measured
GPU board energy during the whole benchmark command. Therefore, the wall-clock
time includes command-level initialization and data/model loading, whereas the
energy and carbon values account only for GPU board electricity.

Operational carbon emissions are estimated as
\begin{equation}
    C_{\mathrm{op}}
    = E_{\mathrm{GPU}}^{\mathrm{kWh}} \times \mathrm{PUE}
    \times I_{\mathrm{grid}},
\label{eq:carbon}
\end{equation}
where $E_{\mathrm{GPU}}^{\mathrm{kWh}}$ denotes the measured GPU board energy
converted to kWh, $\mathrm{PUE}$ is the power usage effectiveness, and
$I_{\mathrm{grid}}$ is the grid carbon intensity. Following a device-level
accounting boundary, we set $\mathrm{PUE}=1.0$ and
$I_{\mathrm{grid}}=475~\mathrm{gCO_2e/kWh}$. Thus, the carbon values should be
interpreted as estimated operational GPU emissions for the measured hardware
and grid setting. They do not include CPU energy, host memory, storage,
networking, data transfer, upstream pretraining, checkpoint production, or
embodied hardware emissions.

Table~\ref{tab:carbon} and Fig.~\ref{fig:carbon_efficiency} show a clear
accuracy--cost trade-off. Note that Table~\ref{tab:main} reports the best testing results to demonstrate the maximum capacity and theoretical upper bound of the models. In contrast, evaluating energy and operational carbon costs requires accounting for hardware-level stochasticity. Thus, these cost metrics are averaged over three independent runs to reflect real-world operational stability. HyperSAM obtains the highest mean OA, AA, and Kappa
among the compared methods, reaching 66.91\% OA, 74.09\% AA, and 62.02\%
Kappa, with an average runtime of 21.11~s, GPU energy of 0.624~Wh, and
estimated emissions of 0.296~g CO$_2$e. Compared with the strongest
non-HyperSAM result in terms of OA, DOFA-LP, HyperSAM improves OA, AA, and
Kappa by 7.59, 9.61, and 6.68 percentage points, respectively, at the cost of
higher GPU energy. HyperFree requires less GPU energy than HyperSAM, but its
classification accuracy is substantially lower. SSFTT and TGRS-ViT occupy the
low-runtime and low-energy region of the plot, yet their OA and Kappa remain
below those of HyperSAM and DOFA-LP.

These results indicate that HyperSAM is not the minimum-energy option. Rather,
it lies at the high-accuracy end of the operational accuracy--energy spectrum.
The additional cost mainly comes from promptable mask proposal generation and
spectral-spatial feature extraction, which provide region-level support
aggregation and improve one-shot classification robustness. The cost should
also be interpreted together with the unified interface evaluated above: once
masks and dense features have been extracted for a scene, they can potentially
be reused for classification, change analysis, anomaly inspection, and target
querying without retraining separate task-specific models. Since the absolute
per-scene GPU emissions remain below one gram of CO$_2$e under the adopted
device-level boundary, the practical deployment decision depends on the
desired accuracy, the number of scenes to be processed, the possibility of
reusing extracted masks and features across tasks, and the carbon intensity of
the actual computing environment.

\begin{table}[t]
\centering
\caption{Accuracy, energy, and operational carbon cost on Indian Pines. Costs are the mean and standard deviation over three cold-start replays of that same configuration and include downstream training when required followed by full-scene inference.}
\label{tab:carbon}
\scriptsize
\setlength{\tabcolsep}{1.8pt}
\resizebox{0.48\textwidth}{!}{
\begin{tabular}{l|ccc|ccc}
\hlinew{1pt}
Method & OA (\%) & AA (\%) & Kappa (\%) &
Time (s) & Energy (Wh) & CO$_2$e (g) \\
\hline
HyperSAM
& \textbf{66.91} & \textbf{74.09} & \textbf{62.02}
& 21.11$\pm$1.98 & 0.624$\pm$0.223 & 0.2964$\pm$0.1059 \\
HyperFree
& 50.36 & 58.06 & 43.62
& 23.40$\pm$18.94 & 0.244$\pm$0.111 & 0.1159$\pm$0.0527 \\
SSFTT
& 54.85 & 65.07 & 49.51
& 8.30$\pm$0.06 & 0.087$\pm$0.001 & 0.0413$\pm$0.0005 \\
TGRS-ViT
& 49.73 & 63.49 & 44.92
& 8.16$\pm$0.13 & 0.120$\pm$0.002 & 0.0570$\pm$0.0010 \\
HyperSIGMA-LP
& 40.69 & 46.99 & 33.53
& 27.01$\pm$37.63 & 0.274$\pm$0.401 & 0.1302$\pm$0.1905 \\
DOFA-LP
& 59.32 & 64.48 & 55.34
& 13.69$\pm$9.88 & 0.181$\pm$0.168 & 0.0860$\pm$0.0798 \\
\hlinew{1pt}
\end{tabular}}
\end{table}

\section{Conclusion}
\label{sec:conclusion}

In this paper, we present \textbf{HyperSAM}, a promptable hyperspectral foundation model that integrates physics-informed data synthesis, SAM3-based spectral adaptation, and confidence-aware robust learning. By reconstructing full-spectrum hyperspectral cubes from SpaceNet multispectral imagery and pairing them with high-resolution SAM3 pseudo-masks, HyperSAM builds object-centric spectral-mask supervision with sharp spatial boundaries and physically plausible spectra. Architecturally, it preserves the geometric and prompt-response priors of the frozen SAM3 RGB branch, injects hyperspectral evidence through a trainable side encoder with zero-initialized residual adapters, and uses a lightweight MoE decoder for scale-adaptive mask refinement. During training, best-mask supervision, CromSS-style confidence weighting, dual spectral-window consistency, and router balance regularization improve robustness to imperfect pseudo-masks. Experiments across HC, HAD, HCD, and HTD show that one trained HyperSAM parameter set can be reused \highlight{under the evaluated small-scene protocols}. The gains over SAM3 verify the \highlight{value} of spectral adaptation \highlight{in this setting}, while the \highlight{comparisons with} hyperspectral baselines demonstrate the \highlight{benefit} of preserving promptable foundation-model priors \highlight{under a common low-label budget}. Ablation results further show that a compact but higher-quality synthetic corpus can be more effective than a larger weakly object-centric corpus. Overall, HyperSAM suggests that reliable object masks, physically plausible spectra, and stable promptable priors are \highlight{useful ingredients for} transferable hyperspectral foundation models. The additional HOSD experiment further demonstrates \highlight{transfer} to airborne hyperspectral oil-spill mapping and environmental monitoring.

\bibliographystyle{IEEEtran}
\bibliography{reference}

\end{document}